# Automated Search for Impossibility Theorems in Social Choice Theory: Ranking Sets of Objects

**Christian Geist**                                                                                       cgeist@gmx.net
**Ulle Endriss**                                                                                     ulle.endriss@uva.nl
*Institute for Logic, Language and Computation*
*University of Amsterdam*
*Postbus 94242*
*1090 GE Amsterdam*
*The Netherlands*

## Abstract

We present a method for using standard techniques from satisfiability checking to automatically verify and discover theorems in an area of economic theory known as *ranking sets of objects*. The key question in this area, which has important applications in social choice theory and decision making under uncertainty, is how to extend an agent's preferences over a number of objects to a preference relation over nonempty sets of such objects. Certain combinations of seemingly natural principles for this kind of preference extension can result in logical inconsistencies, which has led to a number of important impossibility theorems. We first prove a general result that shows that for a wide range of such principles, characterised by their syntactic form when expressed in a many-sorted first-order logic, any impossibility exhibited at a fixed (small) domain size will necessarily extend to the general case. We then show how to formulate candidates for impossibility theorems at a fixed domain size in propositional logic, which in turn enables us to automatically search for (general) impossibility theorems using a SAT solver. When applied to a space of 20 principles for preference extension familiar from the literature, this method yields a total of 84 impossibility theorems, including both known and nontrivial new results.

## 1. Introduction

The area of economic theory known as *ranking sets of objects* (Barberà, Bossert, & Pattanaik, 2004; Kannai & Peleg, 1984) addresses the question of how to extend a preference relation defined over certain individual objects to a preference relation defined over nonempty sets of those objects. This question has important applications. For instance, if an agent is uncertain about the effects of two alternative actions, she may want to rank the relative desirability of the two sets of possible outcomes corresponding to the two actions. In the absence of a probability distribution over possible outcomes, the principles and methods developed in the literature on *ranking sets of objects* can guide this kind of decision making under ("complete") uncertainty (e.g., see Gravel, Marchant, & Sen, 2008; Ben Larbi, Konieczny, & Marquis, 2010). A second example for applications is voting theory. When we want to analyse the incentives of a voter to manipulate an election, i.e., to obtain a better election outcome for herself by misrepresenting her true preferences on the ballot sheet, we need to be able to reason about her preferences in case that election will produce a tie and return a set of winners (e.g., see Gärdenfors, 1976; Duggan & Schwartz, 2000;

Taylor, 2002). In both scenarios, decision making under uncertainty and the manipulation of elections with sets of tied winners, agents are assumed to have preferences over simple "objects" (states of the world, election outcomes) which they then need to extend to sets of such objects to be able to use these extended preferences to guide their decisions.

One line of research has applied the *axiomatic method*, as practised in particular in social choice theory (Gaertner, 2009), to the problem of *ranking sets of objects* and formulated certain principles for extending preferences to set preferences as *axioms*. For instance, the *dominance axiom* states that you should prefer set $A \cup \{x\}$ to set $A$ whenever you prefer the individual object $x$ to any element in $A$ (and $A$ to $A \cup \{x\}$ in case $x$ is worse than any element in $A$); and the *independence axiom* states that if you prefer set $A$ to set $B$, then that preference should not get inverted when we add a new object $x$ to both sets. Against all intuition, Kannai and Peleg (1984) have shown that, for any domain with at least six objects, it is *impossible* to rank sets of objects in a manner that satisfies both dominance and independence. This is the original and seminal result in the field, and since 1984 a small number of additional impossibility theorems have been established.

In this paper we develop a method to automatically search for impossibility theorems like the Kannai-Peleg Theorem, to enable us to both verify the correctness of known results and to discover new ones. There are a number of reasons why this is *useful*. First, the ability to *discover* new theorems is clearly useful whenever those theorems are (potentially) interesting. But also the *verification* of known results has its merits: verification can increase our confidence in the correctness of a result (the manual proof of which may be tedious and prone to errors); verification forces us to fully formalise the problem domain, which will often result in a deeper understanding of its subtleties; and verification of theorems in new fields (here the social and economic sciences) can help advance the discipline of automated reasoning itself by providing new test cases and challenges.

This is not the first time that techniques from logic and automated reasoning have been applied to modelling and verifying results from economic theory. Here we briefly review a number of recent contributions applying such tools to social choice theory, which is closely related to the problem domain we focus on in this paper. A lot of this work has concentrated on Arrow's Theorem, which establishes the impossibility of aggregating preferences of a group of agents in a manner that satisfies certain seemingly natural principles (Arrow, 1963; Gaertner, 2009). Ågotnes, van der Hoek, and Wooldridge (2010), for instance, introduce a modal logic for modelling preferences and their aggregation, while Grandi and Endriss (2009) show that Arrow's Theorem is equivalent to the statement that a certain set of sentences in classical first-order logic does not possess a finite model.

Besides formally modelling the problem domain and the theorem, there have also been a number of attempts to automatically re-prove Arrow's Theorem. One approach has been to encode the individual steps of known proofs into higher-order logic and to then verify the correctness of these proofs with a proof checker. Examples for this line of work are the contributions of Wiedijk (2007), who has formalised a proof of Arrow's Theorem in the MIZAR proof checker, and that of Nipkow (2009), who did the same using the ISABELLE system. A particularly interesting approach is due to Tang and Lin (2009). These authors first prove two lemmas that reduce the general claim of Arrow's Theorem to a statement pertaining to the special case of just *two* agents and *three* alternatives. They then show that this statement can be equivalently modelled as a (large) set of clauses in propositional logic. The inconsistency of this set of clauses can be verified using a SAT solver, which in turn (together with the lemmas) proves the full theorem. Tang and Lin have been able to extend their method to the verification of a number of other results in social choice theory and they have also shown that their method can serve as a useful tool to support the (semi-automatic) testing of hypotheses during the search for new results (Tang & Lin, 2009; Tang, 2010).[1] All of these contributions neatly fit under the broad heading of *computational social choice*, the discipline concerned with the study of the computational aspects of social choice, the application of computational techniques to problems in social choice theory, and the integration of methods from social choice theory into AI and other areas of computer science (Chevaleyre, Endriss, Lang, & Maudet, 2007).

Our starting point for developing a method for automatically proving impossibility theorems in the area of *ranking sets of objects* has been the work of Tang and Lin (2009). We have adapted and extended their approach as follows. Rather than proving a new lemma reducing a general impossibility to an impossibility in a small domain for each and every theorem that we want to verify, our first contribution is a broadly applicable result, the *Preservation Theorem*, which entails that for any combinations of axioms satisfying certain syntactic conditions, any impossibility that can be established for a domain of (small) fixed size $n$ will unravel into a full impossibility theorem for all domains of size $\geq n$. To be able to formulate this result, we introduce a *many-sorted first-order logic* for expressing axioms relating preferences over individual objects to preferences over sets of objects. We were able to express most axioms from the literature in this language, which facilitates a *fully* automated search for impossibility theorems within the space of these axioms.

We then show how impossibility theorems regarding the extension of preferences can be modelled in *propositional logic*, provided the size of the domain is fixed. Given the Preservation Theorem, any inconsistency found with the help of a SAT solver now immediately corresponds to a general impossibility theorem. We have implemented this kind of automated theorem search as a scheduling algorithm that exhaustively searches the space of all potential impossibility theorems for a given set of axioms and a given critical domain size $n$. Together with a number of heuristics for pruning the search space, this approach represents a practical method for verifying existing and discovering new impossibility theorems.

Finally, we have applied our method to a search space defined by 20 of the most important preference extension axioms from the literature, and we have exhaustively searched this space (of around one million possible combinations) for domains of up to eight objects. This search has resulted in 84 (minimal) impossibility theorems. Each theorem states, for a particular $n \leq 8$ and a particular set of axioms $\Delta$, that there exists no preference ordering over nonempty sets of objects that satisfies all the axioms in $\Delta$ when there are $n$ or more objects in the domain. These 84 theorems are minimal in the sense that no strict subset of $\Delta$ would still result in an impossibility (at the given domain size $n$) and in the sense that all the axioms in $\Delta$ can be satisfied for any domain with fewer than $n$ objects.

The 84 impossibility theorems found include known results (such as the Kannai-Peleg Theorem), simple consequences of known results, as well as new and nontrivial theorems that constitute relevant contributions to the literature on *ranking sets of objects*. One of these is the impossibility of combining independence and a weakened form of dominance with two axioms known as *simple uncertainty aversion* and *simple top monotonicity* (see Appendix A). This is particularly interesting, because (in the context of a characterisation of a particular type of set preference orders) this very same set of axioms had previously been claimed to be consistent (Bossert, Pattanaik, & Xu, 2000). This has later been found to be a mistake, which has been corrected by Arlegi (2003), but even his work does not establish the actual impossibility theorem. This certainly demonstrates the nontrivial nature of the problem. Other interesting theorems discovered by our method include variants of the Kannai-Peleg Theorem involving weakened versions of the independence axiom, impossibility theorems that do not rely on the dominance axiom (which is an integral component of most other results in the field), and impossibility theorems for which the critical domain size $n$ is different from those featuring in any of the known results in the literature.

---

1. See the work of Lin (2007) for an outline of their general methodology as well as several examples for applications in domains other than social choice theory.

The remainder of the paper is organised as follows. Section 2 introduces the formal framework of *ranking sets of objects* and recalls the seminal result in the field, the Kannai-Peleg Theorem. In Section 3 we prove our Preservation Theorem, which allows us to reduce general impossibilities to small instances. Section 4 then shows how to model those small instance as sets of clauses in propositional logic. Building on these insights, Section 5 finally presents our method to automatically search for impossibility theorems, as well as the 84 impossibility theorems we have been able to obtain using this method. Section 6 concludes with a brief summary and a discussion of possible directions for future work. Appendix A provides a list of the 20 axioms used in our automated theorem search. The reader can find additional detail, regarding both the method and the impossibility theorems discovered, in the Master's thesis of the first author (Geist, 2010).

## 2. Ranking Sets of Objects

*Ranking sets of objects* deals with the question of how an agent should rank sets of objects, given her preferences over individual objects. Answers to this question will depend on the concrete interpretation assigned to sets (Barberà et al., 2004):

- *Complete uncertainty.* Under this interpretation, sets are considered as containing mutually exclusive alternatives from which the final outcome is selected at a later stage, but the agent does not have any influence on the selection procedure.

- *Opportunity sets.* Here, again, sets contain mutually exclusive alternatives, but this time the agent can pick a final outcome from the set herself.

- *Sets as final outcomes.* In this setting, sets contain compatible objects that are assumed to materialise simultaneously (i.e., the agent will receive all of them together).

Suppose an agent prefers object $x$ over object $y$. Then under the first interpretation it is reasonable to assume that she will rank $\{x\}$ (receiving $x$ with certainty) over $\{x,y\}$ (receiving either $x$ or $y$). Under the second interpretation she might be indifferent between $\{x\}$ and $\{x,y\}$, as she can simply pick $x$ from the latter set. Under the third interpretation, finally, she might prefer $\{x,y\}$, as this will give her $y$ on top of $x$. In this paper, we will focus on the idea of *complete uncertainty*, which is the most studied of the three. An important application of this interpretation is voting theory: when we want to analyse whether a voter has an incentive to manipulate an election we often have to reason about her preferences between alternative outcomes, each producing a set of tied winning candidates (Gärdenfors, 1976; Duggan & Schwartz, 2000; Taylor, 2002).

Next, we introduce the notation and mathematical framework usually employed to treat problems in the field of *ranking sets of objects* (e.g., see Barberà et al., 2004), and we then present the aforementioned Kannai-Peleg Theorem in some detail (Kannai & Peleg, 1984).

### 2.1 Formal Framework

Let $X$ be a (usually finite) set of *alternatives* (or *objects*), on which a (preference) order $\dot{\geq}$ is defined. The order $\dot{\geq}$ is assumed to be *linear*, i.e., it is a reflexive, complete, transitive and antisymmetric binary relation. We denote the strict component of $\dot{\geq}$ by $\dot{>}$, i.e., $x \dot{>} y$ if $x \dot{\geq} y$ and $y \dot{\not\geq} x$. The interpretation of $\dot{\geq}$ will be such that $x \dot{\geq} y$ if and only if $x$ is considered at least as good as $y$ by the decision maker.

Similarly, we have a binary relation $\succeq$ on the set of nonempty subsets of $X$ (denoted by $\mathcal{X} := 2^X \setminus \{\emptyset\}$). This relation will for now be assumed to be a *weak order* (reflexive, complete, transitive); later on, however, our proof method will allow to explore weaker assumptions regarding $\succeq$, too. Like above, we use $\succ$ for the strict component of $\succeq$. Additionally, we also define an indifference relation $\sim$ by setting $A \sim B$ if and only if $A \succeq B$ and $B \succeq A$.

For any $A \in \mathcal{X}$ we write $\max(A)$ for the maximal element in $A$ with respect to $\dot{\geq}$ and $\min(A)$ for the minimal element in $A$ with respect to $\dot{\geq}$.

### 2.2 The Kannai-Peleg Theorem

Kannai and Peleg (1984) were probably the first to treat the specific problem of extending preferences from elements to subsets as a problem in its own right and in an axiomatic fashion. In previous work, other authors had regarded this problem as more of a side issue of other problems, particularly in the analysis of the manipulation of elections (e.g., see Fishburn, 1972; Gärdenfors, 1976), or had merely axiomatised specific methods of extension without considering the general problem (e.g., see Packard, 1979).

The Kannai-Peleg Theorem makes use of two axioms, both of which are very plausible under the interpretation of *complete uncertainty*. First, there is the *Gärdenfors principle* (Gärdenfors, 1976, 1979), also known as *dominance*. This principle consists of two parts and it requires that

(1) adding an element, that is strictly better ($\dot{>}$) than all the elements in a given set, to that given set produces a *strictly* better set with respect to the order $\succeq$,

(2) adding an element, that is strictly worse ($\dot{<}$) than all the elements in a given set, to that given set produces a *strictly* worse set with respect to the order $\succeq$.

Formally, the Gärdenfors principle (`GF`) can be written as the following two axioms:

(`GF1`) $\quad ((\forall a \in A) x \dot{>} a) \Rightarrow A \cup \{x\} \succ A \quad$ for all $x \in X$ and $A \in \mathcal{X}$,

(`GF2`) $\quad ((\forall a \in A) x \dot{<} a) \Rightarrow A \cup \{x\} \prec A \quad$ for all $x \in X$ and $A \in \mathcal{X}$.

Second, we have a monotonicity principle called *independence*, which states that, if a set is strictly better than another one, then adding the same alternative (which was not contained in either of the sets before) to both sets simultaneously does not reverse this strict order. An equivalent way of stating this (in the light of the completeness of the order) is to require that at least a non-strict preference remains of the original strict preference (such that $\succ$ becomes $\succeq$). The formal statement reads as follows:

$$(\text{IND}) \quad A \succ B \Rightarrow A \cup \{x\} \succeq B \cup \{x\} \quad \text{for all } A, B \in \mathcal{X} \text{ and } x \in X \setminus (A \cup B).$$

Before we get to the main theorem, we present a lemma, also due to Kannai and Peleg (1984). It says that only very specific rankings can satisfy the conditions (GF) and (IND).

**Lemma 1.** *If $\succeq$ satisfies the Gärdenfors principle (GF) and independence (IND), then $A \sim \{\max(A), \min(A)\}$ for all $A \in \mathcal{X}$.*

*Proof.* Let $A$ be a nonempty subset of $X$. If $|A| \leq 2$ then the lemma holds trivially by reflexivity of $\succeq$ since then $A = \{\max(A), \min(A)\}$. So suppose $|A| \geq 3$ and define $A_- := A \setminus \max(A)$. Note that, because $|A| \geq 3$, the set $A_-$ is nonempty and thus $\{\min(A)\} = \{\min(A_-)\}$. By a repeated application of (GF1) we get $\{\min(A)\} = \{\min(A_-)\} \prec A_-$. We can then add $\max(A)$ to both sides, showing that $\{\min(A), \max(A)\} \preceq A$ by (IND). In a completely analogous way we get $\{\min(A), \max(A)\} \succeq A_+ \cup \{\min(A)\} = A$ from (GF2) and (IND), where $A_+ := A \setminus \min(A)$. □

That is, Lemma 1 shows that the ranking of subsets is completely determined by their worst and best elements. We are now ready to state and prove the theorem.

**Theorem 1** (Kannai and Peleg, 1984)**.** *Let $X$ be a linearly ordered set with $|X| \geq 6$. Then there exists no weak order $\succeq$ on $\mathcal{X}$ satisfying the Gärdenfors principle (GF) and independence (IND).*

*Proof.* Let $x_i$, $i \in \{1, 2, \ldots, 6\}$ denote six distinct elements of $X$ such that they are ordered by $\dot{>}$ with respect to their index, i.e., $x_1 \dot{>} x_2 \dot{>} x_3 \dot{>} x_4 \dot{>} x_5 \dot{>} x_6$. By way of contradiction, suppose there exists a weak order $\succeq$ on $\mathcal{X}$ satisfying the Gärdenfors principle (GF) and independence (IND). We first claim that then

$$\{x_2, x_5\} \succeq \{x_3\}. \tag{1}$$

In order to prove this claim, suppose that the contrary is the case, which by completeness of $\succeq$ is $\{x_3\} \succ \{x_2, x_5\}$. We can then, by (IND), include $x_6$, which yields $\{x_3, x_6\} \succeq \{x_2, x_5, x_6\}$. Note now that together with Lemma 1 (and transitivity) this implies

$$\{x_3, x_4, x_5, x_6\} \succeq \{x_2, x_3, x_4, x_5, x_6\},$$

contradicting (GF1). Thus, claim (1) must be true and it follows from $\{x_3\} \succ \{x_3, x_4\} \succ \{x_4\}$ (which is a consequence of the Gärdenfors principle) together with transitivity that $\{x_2, x_5\} \succ \{x_4\}$. Using (IND) again, we can add (the so far unused) $x_1$ and get $\{x_1, x_2, x_5\} \succeq \{x_1, x_4\}$. As before, we again fill in the intermediate elements to both sets and obtain, by Lemma 1 and transitivity, that $\{x_1, x_2, x_3, x_4, x_5\} \succeq \{x_1, x_2, x_3, x_4\}$, which this time contradicts (GF2). □

Put differently, the Kannai-Peleg Theorem says that the Gärdenfors principle (GF) and independence (IND) are inconsistent for a domain of six or more elements. Thus, there is no way of extending a linear order on a set of at least six objects to a weak order on the collection of all nonempty sets of these objects. This is why the Kannai-Peleg Theorem is also referred to as an *impossibility theorem*.

Many more axioms are discussed in the literature and we are aware of two more impossibility theorems regarding *choice under complete uncertainty* (Barberà et al., 2004). A selection of 20 of the most important axioms can be found in Appendix A.

## 3. Reduction of Impossibilities to Small Instances

While the Kannai-Peleg Theorem applies to any set $X$ with *at least* six elements, the proof we have given (which closely follows the original proof of Kannai and Peleg) works by exhibiting the case with *exactly* six elements. The fact that the impossibility theorem applies to larger domains as well is very clear in this particular case. Our goal in this section is to prove that this approach can be elevated to a general proof technique: to prove a general impossibility theorem it is sufficient to establish impossibilities for a small instance. Specifically, we will prove what we call the *Preservation Theorem*, which says that certain axioms are preserved in specific substructures. A corollary to this theorem then is a *universal reduction step*, which says that the non-existence of a satisfying relation on a small domain shows that no larger satisfying relation can exist either.[2]

We will work in the framework of mathematical logic in order to have access to the syntactic as well as semantic features of axioms. In Section 3.1, we first describe a many-sorted language for our specific problem of *ranking sets of objects*, before we apply techniques from model theory to prove the Preservation Theorem, in Section 3.2, which has the universal reduction step as a corollary. This universal step is then powerful enough to cater for all the axioms from the literature that we were able to formalise in our language, including all of those listed in Appendix A.

### 3.1 Many-Sorted Logic for Set Preferences

A natural and well-understood language for our problem domain is many-sorted (first-order) logic, which has, compared to first-order logic, more and different quantifiers (allowing for quantification over different domains containing the elements of a respective *sort*), but is still reducible to first-order logic. Apart from the quantifiers, many-sorted logic is practically equivalent to first-order logic and thus many results (e.g., soundness, completeness, compactness, Löwenheim-Skolem properties, etc.) can be transferred from first-order logic or can be directly proven (Manzano, 1996; Enderton, 1972).

Many-sorted logic is characterised by the use of a set $\boldsymbol{S}$ of different *sorts* $s \in \boldsymbol{S}$. A *structure* (or *model*) $\mathfrak{A}$ for many-sorted logic is just like one for first-order logic, but with separate domains $\text{dom}_s(\mathfrak{A})$ for each sort $s \in \boldsymbol{S}$ instead of one single domain. We then have corresponding quantifiers $\forall_s$ and $\exists_s$ for each sort $s$, equipped with the intuitive semantics:

- $\mathfrak{A} \models \forall_s x \varphi(x)$ if and only if $\mathfrak{A} \models \varphi(a)$ for all $a \in \text{dom}_s(\mathfrak{A})$,

- $\mathfrak{A} \models \exists_s x \varphi(x)$ if and only if $\mathfrak{A} \models \varphi(a)$ for some $a \in \text{dom}_s(\mathfrak{A})$.

---

2. The universal reduction step plays a similar role as the *inductive lemmas* of Tang and Lin (2009) play in their work on a computer-aided proof of Arrow's Theorem and other theorems in social choice theory. An important difference is that for the domain of *ranking sets of objects* we are able to prove a *single* such result, which allows us to perform reductions for a wide range of problems, while Tang and Lin had to prove new (albeit similar) lemmas for every new result tackled.

Other than that, many-sorted logic is analogous to first-order logic with the slight difference of having separate variable, function, and relation symbols for the different sorts or combinations of sorts.

In our case, we will have two sorts ($\boldsymbol{S} = \{\varepsilon, \sigma\}$): elements ($\varepsilon$) and sets ($\sigma$). We further demand that there is a relation $\in$ of type $\langle \varepsilon, \sigma \rangle$ as well as two relations $\dot{\geq}$ and $\succeq$ of type $\langle \varepsilon, \varepsilon \rangle$ and $\langle \sigma, \sigma \rangle$, respectively. These will then later be interpreted as the usual membership relation and our linear and weak orders, respectively. But one can have many more relations and functions in the signature, and we will use the following time and again:

- Relations:
  - $\subseteq$, type $\langle \sigma, \sigma \rangle$ (intuitively: *set inclusion*)
  - disjoint, type $\langle \sigma, \sigma \rangle$ (intuitively: true iff sets are disjoint)
  - evencard, type $\langle \sigma \rangle$ (intuitively: true iff the cardinality of a set is even)
  - equalcard, type $\langle \sigma, \sigma \rangle$ (intuitively: true if sets have the same cardinality)

- Functions:
  - $\cup$, type $\langle \sigma, \sigma, \sigma \rangle$ (intuitively: *set union*)
  - $\{\cdot\}$, type $\langle \varepsilon, \sigma \rangle$ (intuitively: transforms an element into the singleton set)
  - replaceInBy, type $\langle \varepsilon, \sigma, \varepsilon \rangle$ (intuitively: replace an element in a set by another element; e.g., $(A \setminus \{a\}) \cup \{b\}$)

We call this language of many-sorted logic (with the two sorts $\varepsilon$ and $\sigma$ and the signature containing exactly the above relations and functions) MSLSP (**M**any-**S**orted **L**ogic for **S**et **P**references). Notation-wise we will sometimes use (the more common) infix notation for certain relations and functions. We will for instance write $A \cup B$, $a \in A$, $A \subseteq B$ and $\{x\}$ instead of $\cup(A, B)$, $\in (a, A)$, $\subseteq (A, B)$ and $\{\cdot\}(x)$, respectively. Furthermore, we will sometimes use negated symbols like $x \notin A$ for $\neg(x \in A)$ as well as the strict relation symbols $A \succ B$ and $x \dot{>} y$ when we mean $A \succeq B \wedge \neg(B \succeq A)$ and $x \dot{\geq} y \wedge \neg(y \dot{\geq} x)$, respectively. Generally, we will use the (standard model-theoretic) notation of Hodges (1997).

MSLSP is expressive enough to formulate many axioms in the literature (including the 20 axioms of Appendix A) and as an example we give the representations of the principle of *independence* and the *Gärdenfors principle* (see Section 2.2):

**Example 1.** (IND) can be formulated in MSLSP:

$$\forall_\sigma A \forall_\sigma B \forall_\varepsilon x \, [(x \notin (A \cup B) \wedge A \succ B) \to A \cup \{x\} \succeq B \cup \{x\}]$$

**Example 2.** (`GF`) can be formulated in MSLSP:

$$\forall_\sigma A \forall_\varepsilon x \left[ (\forall_\varepsilon a(a \in A \to x \mathrel{\dot{\succ}} a)) \to A \cup \{x\} \succ A \right]$$
$$\forall_\sigma A \forall_\varepsilon x \left[ (\forall_\varepsilon a(a \in A \to a \mathrel{\dot{\succ}} x)) \to A \succ A \cup \{x\} \right]$$

Some axioms, however, do not have such straightforward representations in MSLSP. The concept of *weak preference dominance*, proposed (in a slightly stronger form) by Sen (1991), is an example of such an axiom:

(`WPD`) $\left[ (|A| = |B| \text{ and there exists a bijective function } \varphi : A \to B \text{ such that } a \mathrel{\dot{\geq}} \varphi(a) \text{ for all } a \in A) \Rightarrow A \succeq B \right]$ for any two sets $A, B \in \mathcal{X}$.

Even though there is no obvious way to express this axiom in MSLSP, Puppe (1995) showed that (`WPD`) is actually equivalent to an axiom much closer to our formalism, which he calls *preference-basedness* and which is easily seen to be expressible in MSLSP.[3]

(`PB`) $\left[ b \mathrel{\dot{\geq}} a \Rightarrow (A \setminus \{a\}) \cup \{b\} \succeq A \right]$ for all $A \in \mathcal{X}$, $a \in A$, $b \notin A$,

It might, however, be the case that some axioms are not expressible in MSLSP at all. We have, for instance, not been able to translate the axiom of *neutrality* (Nitzan & Pattanaik, 1984; Pattanaik & Peleg, 1984), which says that the manner in which the ranking is lifted from objects to sets of objects does not depend on the names of the objects. This axiom is usually defined in terms of a function $\varphi$ from objects to objects and postulates that whenever $\mathrel{\dot{\geq}}$ is invariant under $\varphi$, then so is $\succeq$.

(`NEU`) $\big[ \big[ \big( x \mathrel{\dot{\geq}} y \iff \varphi(x) \mathrel{\dot{\geq}} \varphi(y) \text{ and } y \mathrel{\dot{\geq}} x \iff \varphi(y) \mathrel{\dot{\geq}} \varphi(x) \big)$
for all $x \in A$, $y \in B] \Rightarrow$
$[A \succeq B \iff \varphi(A) \succeq \varphi(B) \text{ and } B \succeq A \iff \varphi(B) \succeq \varphi(A)]$
for any two sets $A, B \in \mathcal{X}$ and any injective mapping $\varphi : A \cup B \to X$.

### 3.2 Preservation Theorem and Universal Reduction Step

The famous Łoś-Tarski Theorem of classical model theory offers a weak version of the result we are going to prove.[4] It does, however, not cover certain axioms and therefore we have to find a stronger result than what classical model theory can offer. The idea is to be able to preserve a larger class of axioms by making use of our problem-specific features: like, for instance, the element-set framework. Thus, we define the concepts of a *structure for set preferences* as well as *subset-consistent* substructures:

**Definition 1.** An MSLSP-structure $\mathfrak{B}$ is a *structure for set preferences* if it fulfils the following criteria:

---

3. $\forall_\sigma A \forall_\varepsilon a \forall_\varepsilon b \left[ (a \in A \land b \notin A \land b \mathrel{\dot{\geq}} a) \to \text{replaceInBy}(a, A, b) \succeq A \right]$
4. The exact statement of the theorem (for first-order logic) can, for example, be found in Hodges' book (1997) as Corollary 2.4.2 and the proof idea is relatively simple: by contradiction it suffices to show that $\exists_1$-formulas are preserved by embeddings (Hodges, 1997, Thm 2.4.1). The proof of the latter proceeds by induction on the complexity of the formula and the critical case of the existential quantifier does not cause any trouble as witnesses are not "lost" when moving to a larger structure.

1. $\text{dom}_\sigma(\mathfrak{B}) \subseteq 2^{\text{dom}_\varepsilon(\mathfrak{B})}$, i.e., the domain of sort $\sigma$ contains only sets of elements from the domain of sort $\varepsilon$.

2. The relation symbol $\in$ of type $\langle \varepsilon, \sigma \rangle$ is interpreted in its natural way.

If a substructure $\mathfrak{A}$ of a structure for set preferences $\mathfrak{B}$ is a structure for set preferences, too, then it is called a *subset-consistent* substructure.

Note that in a substructure $\mathfrak{A}$ of a structure for set preferences $\mathfrak{B}$ we have

$$\in^{\mathfrak{A}} = \in^{\mathfrak{B}} |_{\text{dom}(\mathfrak{A})},$$

i.e., the symbol $\in$ must be interpreted as the restriction of its interpretation in $\mathfrak{B}$. Hence, it is sufficient to fulfill the first condition for being a *subset-consistent* substructure of $\mathfrak{B}$.

These two semantic conditions suffice for extending the Łoś-Tarski Theorem to a larger class of axioms. But what are the axioms that we can now treat? Let us look at the following (purely syntactic) definition first and then explain our reasons for choosing this particular class.

**Definition 2.** The class of *existentially set-guarded* (ESG) formulas is the smallest class of MSLSP-formulas recursively defined as follows:

- all quantifier-free formulas are ESG,

- if $\psi(\bar{x})$ and $\psi'(\bar{x})$ are ESG, then $\varphi(\bar{x}) := (\psi \wedge \psi')(\bar{x})$ as well as $\varphi'(\bar{x}) := (\psi \vee \psi')(\bar{x})$ are ESG,

- if $\psi(y, \bar{x})$ is ESG, then $\varphi(\bar{x}) := \forall_s y \psi(y, \bar{x})$ is ESG for any sort $s \in \{\varepsilon, \sigma\}$,

- if $\psi(y, \bar{x})$ is ESG, then $\varphi(\bar{x}) := \exists_\varepsilon y (y \in t(\bar{x}) \wedge \psi(y, \bar{x}))$, where $t$ is a term of sort $\sigma$ and $y$ does not occur in $\bar{x}$, is ESG.

The atomic formulas $y \in t(\bar{x})$ of the last condition are called the *set-guards* of the respective quantifiers.

The class of ESG formulas consists of all MSLSP-formulas that only contain set-guarded existential quantifiers $\exists_\varepsilon$ of sort $\varepsilon$, and no existential quantifiers $\exists_\sigma$ of sort $\sigma$ at all.

Note that when we write $\varphi(\bar{x})$, we do not necessarily mean that $\varphi$ contains *all* the variables in the sequence $\bar{x} = (x_0, x_1, x_2, \ldots)$, but just that all (free) variables of $\varphi$ are among those in $\bar{x}$. We will also use the notation $\varphi[\bar{a}]$, with $\bar{a}$ being a sequence of elements, which will mean that the elements $a_0, a_1, a_2, \ldots$ are assigned to the variables $x_0, x_1, x_2, \ldots$.

Intuitively, we will do the following: in axioms we allow existential quantifiers (but only for elements, i.e., of sort $\varepsilon$) as long as they are "guarded" by sub-formulas saying that the respective witness belongs to some set. The sets can also be unions of sets or formed in a different way by the term $t$. The important part is that when moving from a structure to a substructure this set-guard now guarantees that the witness of the existential quantifier is not lost. This is because the witness has to be within a set (as required by the set-guard) that will be situated in the substructure.

Before we explain this further and give the formal proof of this claim, let us look at examples of ESG and non-ESG sentences:

**Example 3.** The axiom (GF1) (and similarly (GF2)) is an ESG sentence:

$$
\begin{aligned}
&& x \mathrel{\dot{\not\succ}} a && \text{(quantifier-free)} \\
& \exists_\varepsilon a (a \in A \wedge && x \mathrel{\dot{\not\succ}} a) && (\text{adding } \exists_\varepsilon) \\
& \exists_\varepsilon a (a \in A \wedge && x \mathrel{\dot{\not\succ}} a) \vee A \cup \{x\} \succ A && (\vee \text{ with quantifier-free}) \\
& \forall_\sigma A \forall_\varepsilon x [\exists_\varepsilon a (a \in A \wedge && x \mathrel{\dot{\not\succ}} a) \vee A \cup \{x\} \succ A] && (\text{adding } \forall_s).
\end{aligned}
$$

Considering the last line of the above example, one can understand why removing elements from $X$ does not affect this axiom. For universal quantifiers a restriction of the domain is no problem anyway. But also the existential witness is not lost: if we suppose it had been removed for some set $A$, then so would have been the set $A$ itself, as the $\sigma$-domain can only contain sets of elements from the $\varepsilon$-domain (by Definition 1). But with $A$ removed there is no need for a witness anymore.

That one cannot just allow arbitrary existential quantifiers without set-guards can be seen when considering the following example, which shows a very simple sentence (with *unguarded* existential quantifiers) that is not preserved in substructures.

**Example 4.** The MSLSP-sentence (axiom)

$$\exists_\varepsilon x \exists_\varepsilon y \exists_\varepsilon z \, [x \neq y \wedge x \neq z \wedge y \neq z],$$

which says that there are at least three distinct elements in the $\varepsilon$-domain of a structure for set preferences, is clearly not preserved in substructures: it holds in all structures for set preferences $\mathfrak{B}$ with at least three elements in $\mathrm{dom}_\varepsilon(\mathfrak{B})$, but fails to hold in any of its substructures $\mathfrak{A}$ with less than three elements in $\mathrm{dom}_\varepsilon(\mathfrak{A})$.

After these examples the reader should have developed some understanding of why ESG sentences are preserved in substructures and why we cannot allow much more. The formal proof of our Preservation Theorem will explain the first part further.

Note that, apart from the case of the existential quantifier, the proof is essentially identical to one direction of the proof of the Łoś-Tarski Theorem for many-sorted logic, which can be carried out on model-theoretic grounds alone. It is just the last part of this proof (the induction step for the existential quantifier) that requires the syntactic restriction (to ESG sentences) as well as the semantic restriction (to subset-consistent substructures), the latter of which we can allow because of our particular problem domain.

**Theorem 2** (Preservation Theorem). *ESG sentences are preserved in subset-consistent substructures, i.e., if $\mathfrak{A}$ is a subset-consistent substructure of a structure for set preferences $\mathfrak{B}$ then $\mathfrak{B} \models \varphi$ implies $\mathfrak{A} \models \varphi$ for any ESG sentence $\varphi$.*

*Proof.* We prove a stronger statement for ESG *formulas* (instead of sentences) by induction on the complexity of the formula:

> *If $\mathfrak{A}$ is a subset-consistent substructure of a structure for set preferences $\mathfrak{B}$ then $\mathfrak{B} \models \varphi[\bar{a}]$ implies $\mathfrak{A} \models \varphi[\bar{a}]$ for any ESG formula $\varphi(\bar{x})$ and any tuple $\bar{a}$ of elements from $\mathrm{dom}(\mathfrak{A})$ (matching the types of $\bar{x}$).*

So let $\mathfrak{B}$ be a structure for set preferences with a subset-consistent substructure $\mathfrak{A}$, let $\varphi(\bar{x})$ be an ESG formula and, furthermore, let $\bar{a}$ be a tuple of elements from $\text{dom}(\mathfrak{A})$ (matching the types of $\bar{x}$).

*Quantifier-free Formulas:* If $\varphi(\bar{x})$ is quantifier-free, a routine but tedious proof leads to the desired results. One has to carry out a few nested inductions on the complexity of terms and formulas, and examples of such proofs can be found in any textbook on Model Theory (e.g., see Hodges, 1997, Theorem 1.3.1). First, one shows by one induction that terms are interpreted in the substructure $\mathfrak{A}$ as they are interpreted in its superstructure $\mathfrak{B}$, i.e.,

$$t^{\mathfrak{A}}[\bar{a}] = t^{\mathfrak{B}}[\bar{a}] \tag{2}$$

for all terms $t(\bar{x})$. This practically immediately follows from the definition of a substructure.

Then one proceeds by another induction proving that atomic formulas hold in $\mathfrak{A}$ if and only if they hold in $\mathfrak{B}$, i.e.,

$$\mathfrak{A} \models \psi[\bar{a}] \iff \mathfrak{B} \models \psi[\bar{a}] \tag{3}$$

for all atomic formulas $\psi(\bar{x})$. As a typical example, suppose that $\varphi(\bar{x})$ is of the form $R(s(\bar{x}), t(\bar{x}))$, where $R$ is a relation symbol and $s(\bar{x})$ as well as $t(\bar{x})$ are terms (matching the type of $R$). Assume $\mathfrak{A} \models R(s[\bar{a}], t[\bar{a}])$, i.e., it holds that $R^{\mathfrak{A}}(s^{\mathfrak{A}}[\bar{a}], t^{\mathfrak{A}}[\bar{a}])$. By (2) this is equivalent to $R^{\mathfrak{A}}(s^{\mathfrak{B}}[\bar{a}], t^{\mathfrak{B}}[\bar{a}])$. Since furthermore $R^{\mathfrak{A}} = R^{\mathfrak{B}}|_{\text{dom}(\mathfrak{A})}$, we even have an equivalence with $R^{\mathfrak{B}}(s^{\mathfrak{B}}[\bar{a}], t^{\mathfrak{B}}[\bar{a}])$, which is just another way of saying $\mathfrak{B} \models R(s[\bar{a}], t[\bar{a}])$.

Finally, one proves the claim for any quantifier-free formula by carrying out induction steps for conjunction $\wedge$, disjunction $\vee$ and negation $\neg$. Note that the step for $\neg$ is why we required both directions in (3).

*Conjunction and Disjunction:* We only show the part for conjunction here; the one for disjunction is completely analogous. If $\varphi(\bar{x})$ is of the form $\psi(\bar{x}) \wedge \psi'(\bar{x})$ and furthermore $\mathfrak{B} \models \varphi[\bar{a}]$, then both $\psi[\bar{a}]$ and $\psi'[\bar{a}]$ must be true in $\mathfrak{B}$. By the induction hypothesis, this carries over to $\mathfrak{A}$ and we get $\mathfrak{A} \models \psi[\bar{a}] \wedge \psi'[\bar{a}]$.

*Universal Quantification:* If $\varphi(\bar{x})$ is of the form $\forall_s y \psi(y, \bar{x})$ with sort $s \in \{\sigma, \varepsilon\}$ and furthermore $\mathfrak{B} \models \varphi[\bar{a}]$, then for all $b$ of sort $s$ in $\text{dom}_s(\mathfrak{B})$ we have that $\mathfrak{B} \models \psi(b, \bar{a})$. Since $\text{dom}_s(\mathfrak{A}) \subseteq \text{dom}_s(\mathfrak{B})$ we can use the induction hypothesis and obtain $\mathfrak{A} \models \psi(b, \bar{a})$ for any $b \in \text{dom}_s(\mathfrak{A})$. This is the same as saying $\mathfrak{A} \models \forall_s y \psi(y, \bar{a})$, i.e., $\mathfrak{A} \models \varphi[\bar{a}]$.

*Existential Quantification:* If $\varphi(\bar{x})$ is of the form $\exists_\varepsilon y[y \in t(\bar{x}) \wedge \psi(y, \bar{x})]$, where $t(\bar{x})$ is a term of sort $\sigma$ and $y$ does not occur in $\bar{x}$, and furthermore $\mathfrak{B} \models \varphi[\bar{a}]$, then there must exist an element $b$ in $\text{dom}_\varepsilon(\mathfrak{B})$ such that

$$\mathfrak{B} \models (y \in t(\bar{x}) \wedge \psi(y, \bar{x}))[b, \bar{a}], \text{ i.e., } \mathfrak{B} \models y \in t(\bar{x})[b, \bar{a}] \text{ and } \mathfrak{B} \models \psi[b, \bar{a}].$$

Hence, if we can show that $b$ is in the $\varepsilon$-domain of $\mathfrak{A}$ and not just of $\mathfrak{B}$, then it follows by the induction hypothesis that also

$$\mathfrak{A} \models \psi[b, \bar{a}],$$

since then $(b, \bar{a})$ is a tuple of elements of $\mathfrak{A}$.

As $\in$ is interpreted naturally in the structure for set preferences $\mathfrak{B}$ and, additionally, $y$ cannot occur in $\bar{x}$, the statement $\mathfrak{B} \models y \in t(\bar{x})[b, \bar{a}]$ boils down to $b \in t^{\mathfrak{B}}[\bar{a}]$, which is equivalent to

$$b \in t^{\mathfrak{A}}[\bar{a}] \tag{4}$$

since $t^{\mathfrak{A}}[\bar{a}] = t^{\mathfrak{B}}[\bar{a}]$, as stated in (2).

The fact that $b$ is an element of $\operatorname{dom}_\varepsilon(\mathfrak{A})$ (and not just of $\operatorname{dom}_\varepsilon(\mathfrak{B})$) is now implied by $t^{\mathfrak{A}}[\bar{a}]$ being in $\operatorname{dom}_\sigma(\mathfrak{A})$, together with $\mathfrak{A}$ being a subset-consistent substructure:

$$\begin{aligned} b \in t^{\mathfrak{A}}[\bar{a}] &\in \operatorname{dom}_\sigma(\mathfrak{A}) \stackrel{(*)}{\subseteq} 2^{\operatorname{dom}_\varepsilon(\mathfrak{A})} \\ \implies b \in t^{\mathfrak{A}}[\bar{a}] &\in 2^{\operatorname{dom}_\varepsilon(\mathfrak{A})} \\ \implies b \in t^{\mathfrak{A}}[\bar{a}] &\subseteq \operatorname{dom}_\varepsilon(\mathfrak{A}), \end{aligned}$$

where $(*)$ marks the point where the subset-consistency of $\mathfrak{A}$ is used.

Hence, we can, as indicated before, apply the induction hypothesis to $\mathfrak{B} \models \psi[b,\bar{a}]$ and obtain $\mathfrak{A} \models \psi[b,\bar{a}]$. Together with $b \in t^{\mathfrak{A}}[\bar{a}]$ it follows that

$$\mathfrak{A} \models \exists_\varepsilon y(y \in t(\bar{x}) \land \psi(y,\bar{x}))[\bar{a}].$$

This way we are done with the proof of the stronger claim (about formulas), which implies the claim of the theorem (about sentences). □

We are now almost ready to apply this theorem to our setting. Note first, however, that the above theorem does not only hold for axioms that are ESG, but also for axioms that are *equivalent* to an ESG sentence in all structures for set preferences (the reason is that they have the same truth value in any such structure). We refer to these axioms as *ESG-equivalent* axioms. In particular this applies to sentences that are *logically equivalent* (i.e., equivalent in *all* structures) to an ESG sentence.

Now, we finally state and prove the corollary applying our general result to the particular problem domain of *ranking sets of objects*.

**Corollary 1** (Universal Reduction Step). *Let $\Gamma$ be a set of ESG (or ESG-equivalent) axioms and let $n \in \mathbb{N}$ be a natural number. If, for any linearly ordered set $Y$ with $n$ elements, there exists no binary relation on $\mathcal{Y} = 2^Y \setminus \{\emptyset\}$ satisfying $\Gamma$, then also for any linearly ordered set $X$ with more than $n$ elements there is no binary relation on $\mathcal{X} = 2^X \setminus \{\emptyset\}$ that satisfies $\Gamma$.*

*Proof.* Let $\Gamma$ be a set of ESG (or ESG-equivalent) axioms and let $n \in \mathbb{N}$ be a natural number. Assume that for any linearly ordered set $Y$ with $n$ elements, there exists no binary relation on $\mathcal{Y} = 2^Y \setminus \{\emptyset\}$ satisfying $\Gamma$. By way of contradiction, suppose $X$ is a linearly ordered set with $|X| > n$ and there is a binary relation on $\mathcal{X} = 2^X \setminus \{\emptyset\}$ that satisfies $\Gamma$. We can view $X \cup \mathcal{X}$ as a structure for set preferences and define a subset-consistent substructure by restricting $X \cup \mathcal{X}$ to a domain $Y \cup \mathcal{Y}$ with $Y \subseteq X$, $|Y| = n$ and $\mathcal{Y} := 2^Y \setminus \{\emptyset\}$. By the Preservation Theorem all ESG(-equivalent) axioms are preserved in this subset-consistent substructure $Y \cup \mathcal{Y}$. Hence, $Y$ must be a linearly ordered set (as all order axioms are ESG) and, furthermore, there is a binary relation on $\mathcal{Y}$ satisfying $\Gamma$. Contradiction! □

*Remark.* When we say "a given set of ESG axioms", then this deserves some further explanation. We mean axioms that are ESG in MSLSP. One can consider additional relations and functions to be added to this signature, but this holds some hidden challenges. For instance, it is not possible to include a predicate *isWholeSet*, which is true of the whole domain only, a function $(\cdot)^c$ for the complement, or even just the constant symbol $\dot{X}$ referring to the whole domain, since all three would (in their natural interpretation) prevent $Y \cup \mathcal{Y}$ from being a substructure: for example, for any $Y \subset X$, while isWholeSet($Y$) (or equivalently, $Y = \dot{X}$) is false in $X \cup \mathcal{X}$, it is true in $Y \cup \mathcal{Y}$. Similarly, we run into problems when including functions like $\cap$, $\setminus$ or $(\cdot)^c$, which are (in their natural interpretation) not functions in the strict sense in a structure like $X \cup \mathcal{X}$ as they can produce the empty set, which is not in $\mathcal{X}$. Therefore, some attention has to be paid when adding new relation or function symbols to the language in order to capture more axioms.

On the basis of Corollary 1 we can finally do what we had been hoping for: in order to prove new impossibility theorems and check existing ones, we only have to look at their base cases (as long as all axioms involved are expressible in MSLSP and are ESG-equivalent). As we shall see next, these small instances can be efficiently checked on a computer.

## 4. Representing Small Instances in Propositional Logic

The small instances, which we can reduce impossibilities to, now need to be checked on a computer. This requires a clever approach as a direct check is far too expensive. We therefore modify and extend a technique due to Tang and Lin (2009).

It is remarkable that Tang and Lin (2009) were able to formulate the base case of Arrow's Theorem in propositional logic, even though some of the axioms intuitively are second-order statements. The trick they used was to introduce "situations" as names for preference profiles, which transforms the second-order axioms into first-order statements, which can then (because of the finiteness of the base case) be translated into propositional logic. We are going to use a similar approach since the axioms for *ranking sets of objects* are stated in a (somewhat enriched)[5] second-order format, too. The setting of *ranking sets of objects* will, however, require a different treatment (which we are going to discuss in the sequel) since we also have to apply functions like union ($\cup$) and singleton set ($\{\cdot\}$) to sets and elements, respectively, whereas no functions needed to be applied to Tang and Lin's situations. Instead of coding these operations on sets *within* the propositional language, we let the program that generates the final formula handle them.

In this section, we first show how to translate axioms for *ranking sets of objects* into propositional logic and then explain how to instantiate instances of these axioms for fixed domain sizes on a computer.

### 4.1 Conversion to Propositional Logic

As an example, consider again the Kannai-Peleg Theorem (Theorem 1). In light of the universal reduction step (Corollary 1), proving this theorem reduces to proving a small base case of exactly six elements:

**Lemma 2** (Base case of the Kannai-Peleg Theorem). *Let $X$ be a linearly ordered set with exactly 6 elements. Then there exists no weak order $\succeq$ on $\mathcal{X}$ satisfying the Gärdenfors principle* (GF) *and independence* (IND).

It might seem tempting to perform a direct check of the involved axioms on all weak orders over the nonempty subsets of a six-element space. This, however, can be seen to be practically impossible as there are around $1.525 \cdot 10^{97}$ such orderings (Sloane, 2010, integer

---

5. There is also an order (i.e., a relation) on sets.

156

sequence A000670). Therefore, we should stick to the idea of transforming the axioms of the Kannai-Peleg Theorem into propositional logic such that they can be checked by a SAT solver, which usually operates on propositional formulas in conjunctive normal form (CNF) only. We will describe in the following how instances of axioms, like the ones stated in Appendix A, can be converted to that language.

It will be sufficient for our formalisation to have two kinds of propositions only: $w(A, B)$ and $l(x, y)$ (corresponding to propositional variables $w_{A,B}$ and $l_{x,y}$) with intended meanings *A is ranked at least as high as B by the weak order $\succeq$* (or short: $A \succeq B$), and *x is ranked at least as high as y by the linear order $\dot{\geq}$* (or short: $x \dot{\geq} y$), respectively. For example, for the base case of the Kannai-Peleg Theorem this leads to a maximum of $|\mathcal{X}|^2 + |X|^2 = (2^6 - 1)^2 + 6^2 = 4005$ different propositional variables.

As indicated earlier, the axioms for *linear* and *weak orders* on $X$ and $\mathcal{X}$, respectively, are entirely unproblematic as they only contain first-order quantifications and, thus, they can easily be transformed. As an example, we include here the transformation of the transitivity axiom for orders on sets:

$$(\text{TRANS}_\sigma) \quad (\forall A \in \mathcal{X})(\forall B \in \mathcal{X})(\forall C \in \mathcal{X}) [A \succeq B \wedge B \succeq C \to A \succeq C]$$
$$\cong \bigwedge_{A \in \mathcal{X}} \bigwedge_{B \in \mathcal{X}} \bigwedge_{C \in \mathcal{X}} [\neg w_{A,B} \vee \neg w_{B,C} \vee w_{A,C}].$$

Note, that, due to the finiteness of $X$ (and thus $\mathcal{X}$), the derived formulas are actually finite objects and can therefore be instantiated by hand (requiring a lot of effort) or using a computer. Furthermore, only very little or no work is needed to convert them into CNF.

Other axioms, like (GF) and (IND), appear to be more difficult to transform because of functions like singleton set $\{\cdot\} : X \to \mathcal{X}$ and set union $\cup : \mathcal{X} \times \mathcal{X} \to \mathcal{X}$, which occur within these axioms. In fact, however, the same simple conversion technique as above can be applied since we are going to take care of those (and similar) functions automatically in our computer program for the instantiation of the axioms. How we do this will be briefly described in Section 4.2 and for now we treat terms like $A \cup B$ as if they were just the corresponding objects from the functions range, i.e., images under the respective functions. For example, this leads to the following conversion of (GF1):

$$(\text{GF1}) \quad (\forall A \in \mathcal{X})(\forall x \in X)[((\forall a \in A) x \dot{>} a) \to A \cup \{x\} \succ A]$$
$$\cong \bigwedge_{A \in \mathcal{X}} \bigwedge_{x \in X} \left[ \left( \bigwedge_{a \in A} l_{x,a} \wedge \neg l_{a,x} \right) \to (w_{A \cup \{x\}, A} \wedge \neg w_{A, A \cup \{x\}}) \right]$$
$$\equiv \bigwedge_{A \in \mathcal{X}} \bigwedge_{x \in X} \left[ \left( \left( \bigvee_{a \in A} \neg l_{x,a} \vee l_{a,x} \right) \vee w_{A \cup \{x\}, A} \right) \wedge \left( \left( \bigvee_{a \in A} \neg l_{x,a} \vee l_{a,x} \right) \vee \neg w_{A, A \cup \{x\}} \right) \right],$$

where the last step only serves the purpose of converting into CNF.

The remaining problematic parts of the formula are the propositional variable index $A \cup \{x\}$, and the disjunction domain criterion $a \in A$. In order to write out the formula explicitly (which we need to do to be able to feed it to a SAT solver) we have to determine which set is represented by $A \cup \{x\}$ and also decide whether $a \in A$ for any $a \in X$. In different words, what we need is explicit access to the elements of a set and also we have to be able to manipulate them. This would theoretically be possible by hand; practically, however, the instantiation of the formula is far too large to be written out manually. Therefore, we need a computer program for the final conversion step, which we are going to describe in the following section.

As a second example, consider the axiom of independence (IND), which can also be transformed in the above fashion by first using finiteness to replace the quantifiers, and then normalising the formula into CNF:

$$
\begin{aligned}
\text{(IND)} \quad & (\forall A, B \in \mathcal{X})(\forall x \in X \setminus (A \cup B)) \left[ A \succ B \to A \cup \{x\} \succeq B \cup \{x\} \right] \\
\cong \; & \bigwedge_{A,B \in \mathcal{X}} \bigwedge_{\substack{x \in X \\ x \notin (A \cup B)}} \left[ (w_{A,B} \wedge \neg w_{B,A}) \to w_{A \cup \{x\}, B \cup \{x\}} \right] \\
\equiv \; & \bigwedge_{A,B \in \mathcal{X}} \bigwedge_{\substack{x \in X \\ x \notin (A \cup B)}} \left[ \neg w_{A,B} \vee w_{B,A} \vee w_{A \cup \{x\}, B \cup \{x\}} \right].
\end{aligned}
$$

The problematic terms here are $x \notin (A \cup B)$, $A \cup \{x\}$ and $B \cup \{x\}$. But, as we will see, they can —just like the critical terms mentioned before— be handled by our program.

The same method of translation easily extends to other axioms, in particular all those listed in Appendix A (Geist, 2010).

### 4.2 Instantiation of the Axioms on a Computer

As indicated above, we make use of a computer program in order to write out the formulas derived above explicitly. We now briefly discuss the ideas of our implementation and, in particular, the methods employed to cater for previously problematic expressions, such as $A \cup \{x\}$ and $a \in A$. Full details of the implementation are given by Geist (2010).

The most widely used SAT solvers work on input files written according to the DIMACS CNF format (DIMACS, 1993). In a few words, this format requires the propositional variables to be represented by natural numbers (starting from 1, since 0 is used as a separator) with a minus (-) in front for negated literals. Furthermore, the whole file needs to be in CNF; it has to contain exactly one clause per line.

To achieve a formulation of the axioms in this target format, the main idea is to fix an enumeration of all propositional variables (of type $l_{x,y}$ and $w_{A,B}$ with $x, y \in X$ and $A, B \in \mathcal{X}$) by first enumerating sets and elements, and subsequently combining pairs of these using a pairing function. Functions and relations like *union* and *element of* can then be defined to operate on numbers directly, and quantifiers can be translated to conjunctive or disjunctive iterations over their respective domains. All in all, easily readable code can be used to instantiate the axioms.

Since the numberings of the items under consideration (here: elements, sets and later propositional variables) form the core of our implementation, we start the translation process by first fixing an (arbitrary) numbering of the $n$ elements $x \in X$, i.e., a bijective function $c_n : X \to \{0, 1, \ldots, n-1\}$.[6] For the Kannai-Peleg Theorem with its six elements, for instance, the codes will hence range from 0 to 5.

---

6. In contrast to the propositional variables, there are no numbering constraints for the elements and so their numbers are allowed to start from 0.

We can then specify the corresponding numbering of sets in $\mathcal{X}$. This requires special attention because we want to define it in such a way that treating the problematic terms (as mentioned above) is as easy as possible. A natural way to do this is by looking at a set as its characteristic function and converting the corresponding finite string of zeros and ones to a natural number. This allows us to perform operations on the codes of sets directly and hence it is straightforward to instantiate the formulas from Appendix A automatically. All that needs to be done is translating them to our specific source code style. Quantifiers then correspond to *for*-loops over all elements or sets, respectively, and restrictions of the quantification domain as well as operations on elements and sets can be taken care of by functions operating on the "codes" of sets and elements directly.

## 5. Automated and Exhaustive Theorem Search

In Section 4 we described how to generate a (long) formula in propositional logic representing small instances of impossibility theorems for *ranking sets of objects*. Let us denote such a formula by $\varphi$. The formula $\varphi$ describes a model with a linear order on its universe of a given number of elements and a weak order satisfying a given set of axioms on the set of the nonempty subsets of its universe. If such a model exists, $\varphi$ has a satisfying assignment (the explicit description of both orders) and, thus, a (complete) SAT solver will discover this (assuming that there are no time or memory bounds). If, conversely, such a model does not exist —which is exactly the statement of the impossibility theorem— then $\varphi$ is unsatisfiable and, again, a SAT solver will be able to detect this (assuming, again, that there are no time or memory bounds).

By our universal reduction step (Corollary 1), a full impossibility theorem is therefore equivalent to a lemma of the form "the formula $\varphi$ is unsatisfiable".

For example, feeding $\varphi_{\text{KP}}$ (a description of the base case of the Kannai-Peleg Theorem as generated by our program) to the SAT solver zChaff (SAT Research Group, Princeton University, 2007) returns the correct result ('UNSAT') in about 5 seconds and thus the automatic verification of this theorem is complete.

Using this technique for single theorems is likely to produce good results and might be a helpful tool from a practical perspective, but we can do more: we are now going to present a method for a fully automated and exhaustive theorem search for impossibility theorems. Our theorem search will, for a given set of axioms, systematically check which of its subsets are inconsistent and from which smallest domain size onwards these impossibilities do occur, thereby automatically identifying all impossibility theorems that the given axioms can produce.[7] In this sense, the search method is exhaustive on the space of given axioms.

To test the fruitfulness of this approach we ran the search on a set of 20 axioms from the literature (Barberà et al., 2004), which we list and describe in Appendix A. Our search algorithm returned a total of 84 impossibilities, of which a few were known already (and are hence now automatically verified), others are immediate consequences of known results, while others again are surprising and new. The very same search method can also be run with arbitrary other ESG axioms in the field of *ranking sets of objects*, including, for instance, the ones of the interpretation of sets as *opportunity sets* from which the decision maker can himself select his favourite outcome, or, similarly, with axioms for the case of assuming that the agent receives the whole set of alternatives (see Section 2).

---

7. Since for practical reasons we can only check base cases up to a certain domain size $|X| = n$, there could theoretically be more impossibilities hidden that only occur from larger domain sizes onwards.

In the remainder of this section, we first describe our method for automated theorem search and then list and discuss the impossibilities that were found.

### 5.1 Approach

Our search method systematically decides whether combinations of given axioms are compatible or incompatible. We will therefore in the following refer to axiom subsets as *problems*, and for a particular domain size we will speak of a *problem instance*.

After the generation of a problem instance by our computer program (as described in Section 4.2) the instance is then passed to a SAT solver, which returns whether it is a "possible" or an "impossible" one. We implemented interfaces for the commonly used solvers PrecoSAT (Biere, 2010) and zChaff (SAT Research Group, Princeton University, 2007). The latter provides an additional layer of verification by generating a proof trace that can be checked using external tools, while the former is usually faster in practice, but does not have this extra feature.

We could now just run this program on all possible problem instances from a given set of axioms and a maximal domain size individually and collect the results. Note, however, that on a space of 20 axioms with a maximal domain size of eight elements, we already have to deal with a total of $(2^{20} - 1) \cdot 8 \approx 8,400,000$ problem instances. If each of them just requires a running time of one second,[8] the whole job would take roughly 100 days. Therefore, we designed a scheduler that makes sure all axiom subsets are treated for all domain sizes in a sensible order. The order in which we check the problem instances has a big effect on the overall running time because one can make use of a combination of four different effects:

(1) if a set of axioms is *inconsistent* at domain size $|X| = n$, then it will also be *inconsistent* for all *larger* domain sizes $|X| > n$ (Corollary 1),

(2) if a set of axioms is *inconsistent* at domain size $|X| = n$, then also all its (axiom) *supersets* are *inconsistent* at this domain size $|X| = n$,

(3) if a set of axioms is *consistent* at domain size $|X| = n$, then it will also be *consistent* for all *smaller* domain sizes $|X| < n$. (Theorem 2), and

(4) if a set of axioms is *consistent* at domain size $|X| = n$, then also all its (axiom) *subsets* are *consistent* at this domain size $|X| = n$.

Since larger instances require exponentially more time (there are exponentially more variables in the satisfiability problem due to exponentially more subsets in $\mathcal{X}$), we start our search at the smallest domain size and then after completely solving this "level" move on to the next domain size.[9] On a new level, only problems have to be considered that still have the status "possible" because of condition (1) above.

---

8. In our tests, especially larger instances required much more time to be solved on average.
9. It is for this reason that condition (3) is not very helpful in practice.

On each level, as soon as we find an impossibility, we can, by condition (2), mark all axiom supersets as impossible at the current domain size (if they had not been found to be impossible at a smaller domain size already). In order to use this mechanism as efficiently as possible, we must check small axiom sets first. But also the dual approach of starting from large axiom sets and marking all axiom subsets as "compatible" as soon as we find a possibility (condition 4), is an option. In experiments, we found that the best performance is achieved when combining these two approaches and so we decided to run the search in alternating directions (switching every 15 minutes):[10] from large axiom sets to small ones and the other way around.

From a practical point of view, our implementation comes with the limitations of only being able to treat at most 21 axioms at the same time (stack overflows occurred for larger axiom sets) and at most a domain size of eight elements (due to memory limits in the SAT solvers). But with better memory management and improved versions of the SAT solvers, these (practical) boundaries should be extendable further.

### 5.2 Results

Our theorem search (checking problem instances up to a domain size of eight) yields a total of 84 *minimal* impossibility theorems on the space of the 20 selected axioms. The results are minimal in two senses:

- the corresponding axiom set is minimal with respect to set inclusion, i.e., all proper subsets are compatible at the given domain size; and

- the domain size is minimal, i.e., for all smaller domain sizes the given axiom set is still compatible.

Counting the total number of incompatible axiom sets (i.e., including all supersets), we find 312,432 inconsistent axiom sets out of about one million possible combinations.

The whole experiment required a running time of roughly one day for handling all of the nearly 8.5 million instances.[11] In order to externally verify as many of the impossibilities as possible, we used the solver zChaff, which can create a computer-verifiable proof trace, for all instances up to domain size 7, and switched to the faster solver PrecoSAT, which does not have this feature, for instances with (exponentially larger) domain size 8.[12]

In Table 1 we list all minimal impossibilities that our search method was able to find (and hence all that there are) for domain sizes up to 8. Recall again that, by Corollary 1, these all directly correspond to full impossibility results (from the given domain size upwards). The results are presented in ascending order by minimal domain size, and in ascending order by the number of axioms involved as a second criterion, so as to have stronger and easier to grasp impossibilities higher up in table. The axioms and their abbreviated names are listed in Appendix A.

---

10. Switching every 15 minutes turned out to result in good performance, but we have not attempted to systematically optimise this parameter.
11. The experiment was performed on an Intel Xeon 2,26 GHz octo-core machine using only one core and 5GB of the available 24GB memory. The machine is part of the Dutch national compute cluster *Lisa*.
12. The five impossibilities occurring only from domain size 8 onwards have therefore not been verified externally. Using zChaff for these instances (and subsequently verifying them) would have also been possible, but slower by about a factor of 10.

| No. | Size | LIN$_\varepsilon$ | REFL$_\sigma$ | COMPL$_\sigma$ | TRANS$_\sigma$ | EXT | SDom | GF1 | GF2 | IND | strictIND | SUAv | SUAp | STopMon | SBotMon | topIND | botIND | disIND | intIND | evenExt | MC |
|---|---|---|---|---|---|---|---|---|---|---|---|---|---|---|---|---|---|---|---|---|---|
| 1 | 3 | ✓ | · | · | · | · | ✓ | · | · | · | ✓ | · | · | · | · | · | · | · | · | · | · |
| 2 | 3 | ✓ | · | · | · | · | · | · | · | · | · | ✓ | ✓ | · | · | · | · | · | · | · | · |
| 3 | 3 | ✓ | · | · | · | · | · | ✓ | ✓ | · | ✓ | · | · | · | · | · | · | · | · | · | · |
| 4 | 3 | ✓ | · | ✓ | · | · | · | · | ✓ | · | ✓ | · | · | · | · | · | · | · | · | · | ✓ |
| 5 | 3 | ✓ | · | · | · | ✓ | · | · | ✓ | · | ✓ | · | · | · | · | · | · | · | · | · | ✓ |
| 6 | 3 | ✓ | · | · | · | · | · | · | ✓ | · | ✓ | ✓ | · | · | · | · | · | · | · | · | ✓ |
| 7 | 3 | ✓ | · | · | · | · | · | · | ✓ | · | ✓ | · | · | · | · | ✓ | · | · | · | · | ✓ |
| 8 | 4 | ✓ | · | · | · | · | · | · | ✓ | · | ✓ | · | · | · | · | ✓ | · | · | · | · | ✓ |
| 9 | 4 | ✓ | · | · | ✓ | · | · | ✓ | ✓ | ✓ | · | ✓ | · | · | · | · | · | · | · | · | · |
| 10 | 4 | ✓ | · | · | ✓ | · | ✓ | ✓ | · | ✓ | · | ✓ | · | · | · | · | · | · | · | · | · |
| 11 | 4 | ✓ | · | · | ✓ | ✓ | · | ✓ | · | · | ✓ | ✓ | · | · | · | · | · | · | · | · | · |
| 12 | 4 | ✓ | · | · | ✓ | · | · | ✓ | ✓ | ✓ | · | · | ✓ | · | · | · | · | · | · | · | · |
| 13 | 4 | ✓ | · | · | ✓ | · | ✓ | · | ✓ | ✓ | · | · | ✓ | · | · | · | · | · | · | · | · |
| 14 | 4 | ✓ | · | · | ✓ | ✓ | · | · | ✓ | · | · | ✓ | · | ✓ | · | · | · | · | · | · | · |
| 15 | 4 | ✓ | · | · | ✓ | · | · | ✓ | · | ✓ | · | ✓ | · | ✓ | · | · | · | · | · | · | · |
| 16 | 4 | ✓ | · | · | ✓ | · | ✓ | · | · | ✓ | · | ✓ | · | ✓ | · | · | · | · | · | · | · |
| 17 | 4 | ✓ | · | · | ✓ | · | · | ✓ | · | · | ✓ | ✓ | · | ✓ | · | · | · | · | · | · | · |
| 18 | 4 | ✓ | · | · | ✓ | · | · | · | ✓ | ✓ | · | · | ✓ | · | · | ✓ | · | · | · | · | · |
| 19 | 4 | ✓ | · | · | ✓ | · | ✓ | · | · | ✓ | · | · | ✓ | · | · | ✓ | · | · | · | · | · |
| 20 | 4 | ✓ | · | · | ✓ | · | · | ✓ | · | ✓ | · | · | ✓ | · | · | ✓ | · | · | · | · | · |
| 21 | 4 | ✓ | · | · | ✓ | ✓ | · | · | · | ✓ | ✓ | · | · | · | · | · | · | · | · | · | ✓ |
| 22 | 4 | ✓ | · | · | ✓ | · | · | ✓ | ✓ | · | · | · | ✓ | · | · | · | · | · | · | · | ✓ |
| 23 | 4 | ✓ | · | · | ✓ | · | · | ✓ | · | · | ✓ | · | ✓ | · | · | · | · | · | · | · | ✓ |
| 24 | 4 | ✓ | · | · | ✓ | · | · | · | · | · | ✓ | · | ✓ | · | · | ✓ | · | · | · | · | ✓ |
| 25 | 4 | ✓ | · | · | ✓ | · | · | · | · | · | · | ✓ | ✓ | · | · | ✓ | · | · | · | · | ✓ |
| 26 | 4 | ✓ | · | · | ✓ | · | · | · | · | · | · | · | ✓ | · | · | ✓ | ✓ | · | · | · | ✓ |
| 27 | 4 | ✓ | · | · | ✓ | · | · | · | · | · | · | · | ✓ | · | · | ✓ | · | ✓ | · | · | ✓ |
| 28 | 4 | ✓ | · | · | ✓ | · | · | ✓ | · | · | ✓ | · | · | · | · | ✓ | ✓ | · | · | · | · |
| 29 | 4 | ✓ | · | · | ✓ | · | ✓ | ✓ | · | · | ✓ | · | · | · | · | ✓ | ✓ | · | · | · | · |
| 30 | 4 | ✓ | · | · | ✓ | · | · | ✓ | ✓ | · | · | · | · | ✓ | · | ✓ | ✓ | · | · | · | · |
| 31 | 4 | ✓ | · | · | ✓ | · | ✓ | · | ✓ | · | · | · | · | ✓ | · | ✓ | ✓ | · | · | · | · |
| 32 | 4 | ✓ | · | · | ✓ | · | · | ✓ | · | · | · | ✓ | · | ✓ | · | ✓ | ✓ | · | · | · | · |
| 33 | 4 | ✓ | · | · | ✓ | · | ✓ | · | · | · | ✓ | · | · | ✓ | · | ✓ | ✓ | · | · | · | · |
| 34 | 4 | ✓ | · | · | ✓ | · | · | · | ✓ | · | · | ✓ | · | ✓ | ✓ | ✓ | · | · | · | · | · |
| 35 | 4 | ✓ | · | · | ✓ | · | ✓ | · | · | · | · | ✓ | · | ✓ | ✓ | ✓ | · | · | · | · | · |
| 36 | 4 | ✓ | · | · | ✓ | · | · | · | ✓ | · | · | ✓ | · | ✓ | · | · | ✓ | ✓ | · | · | · |
| 37 | 4 | ✓ | · | · | ✓ | · | ✓ | · | · | · | · | ✓ | · | ✓ | · | · | ✓ | ✓ | · | · | · |
| 38 | 4 | ✓ | · | · | ✓ | · | · | ✓ | · | · | · | ✓ | · | · | ✓ | · | ✓ | · | ✓ | · | · |
| 39 | 4 | ✓ | · | · | ✓ | · | ✓ | · | · | · | · | ✓ | · | · | ✓ | · | ✓ | · | ✓ | · | · |
| 40 | 4 | ✓ | · | · | ✓ | · | · | ✓ | ✓ | · | · | ✓ | · | · | · | · | ✓ | · | · | · | ✓ |
| 41 | 4 | ✓ | · | · | ✓ | · | ✓ | ✓ | · | · | · | ✓ | · | · | · | · | ✓ | · | · | · | ✓ |
| 42 | 4 | ✓ | · | · | ✓ | · | · | · | ✓ | · | · | ✓ | · | · | · | ✓ | ✓ | · | · | · | ✓ |

Table 1: Results of our automated and exhaustive theorem search on a space of 20 axioms (including orders).

| No. | Size | LIN$_\varepsilon$ | REFL$_\sigma$ | COMPL$_\sigma$ | TRANS$_\sigma$ | EXT | SDom | GF1 | GF2 | IND | strictIND | SUAv | SUAp | STopMon | SBotMon | topIND | botIND | disIND | intIND | evenExt | MC |
|---|---|---|---|---|---|---|---|---|---|---|---|---|---|---|---|---|---|---|---|---|---|
| 43 | 4 | ✓ | · | · | ✓ | · | · | · | ✓ | · | · | · | ✓ | · | · | · | ✓ | ✓ | · | · | ✓ |
| 44 | 5 | ✓ | · | · | ✓ | · | · | ✓ | · | ✓ | · | ✓ | · | · | · | · | · | · | · | · | · |
| 45 | 5 | ✓ | · | · | ✓ | · | · | ✓ | · | · | ✓ | ✓ | · | · | · | · | · | · | · | · | · |
| 46 | 5 | ✓ | · | · | ✓ | · | · | · | ✓ | ✓ | · | ✓ | · | · | · | · | · | · | · | · | · |
| 47 | 5 | ✓ | · | · | ✓ | · | · | · | ✓ | · | ✓ | · | ✓ | · | · | · | · | · | · | · | · |
| 48 | 5 | ✓ | · | · | ✓ | · | · | ✓ | · | · | · | ✓ | · | · | · | ✓ | ✓ | · | · | · | · |
| 49 | 5 | ✓ | · | · | ✓ | · | · | · | ✓ | · | · | ✓ | · | · | · | ✓ | ✓ | · | · | · | · |
| 50 | 5 | ✓ | · | · | ✓ | · | · | · | ✓ | · | · | ✓ | · | · | · | · | ✓ | ✓ | · | · | · |
| 51 | 5 | ✓ | · | · | ✓ | · | · | ✓ | · | · | · | ✓ | · | · | · | ✓ | · | ✓ | · | · | · |
| 52 | 5 | ✓ | · | ✓ | ✓ | · | · | · | · | · | ✓ | ✓ | · | · | · | · | · | · | · | · | ✓ |
| 53 | 5 | ✓ | · | · | ✓ | · | · | · | · | · | ✓ | ✓ | · | · | ✓ | · | · | · | · | · | ✓ |
| 54 | 5 | ✓ | · | · | ✓ | · | · | ✓ | · | · | · | ✓ | · | · | · | · | ✓ | · | · | · | ✓ |
| 55 | 5 | ✓ | · | · | ✓ | · | · | ✓ | · | · | · | ✓ | · | · | · | · | · | ✓ | · | · | ✓ |
| 56 | 6 | ✓ | · | · | ✓ | · | · | · | · | · | ✓ | ✓ | · | · | · | · | · | · | · | · | ✓ |
| 57 | 6 | ✓ | · | ✓ | ✓ | · | · | ✓ | ✓ | ✓ | · | · | · | · | · | · | · | · | · | · | · |
| 58 | 6 | ✓ | · | ✓ | ✓ | · | ✓ | · | ✓ | ✓ | · | · | · | ✓ | · | · | · | · | · | · | · |
| 59 | 6 | ✓ | · | ✓ | ✓ | · | ✓ | ✓ | · | ✓ | · | · | · | · | · | ✓ | · | · | · | · | · |
| 60 | 6 | ✓ | · | ✓ | ✓ | · | ✓ | · | · | ✓ | · | · | · | · | · | ✓ | ✓ | · | · | · | · |
| 61 | 6 | ✓ | · | ✓ | ✓ | · | · | ✓ | ✓ | · | · | · | · | · | · | ✓ | ✓ | · | · | · | · |
| 62 | 6 | ✓ | · | · | ✓ | · | · | ✓ | ✓ | · | · | ✓ | · | · | · | · | ✓ | · | ✓ | · | · |
| 63 | 6 | ✓ | · | · | ✓ | · | ✓ | ✓ | · | · | · | ✓ | · | · | · | · | ✓ | · | ✓ | · | · |
| 64 | 6 | ✓ | · | · | ✓ | · | · | ✓ | · | · | · | ✓ | · | · | ✓ | · | ✓ | · | ✓ | · | · |
| 65 | 6 | ✓ | · | · | ✓ | · | · | ✓ | ✓ | · | · | · | ✓ | · | · | ✓ | · | · | ✓ | · | · |
| 66 | 6 | ✓ | · | · | ✓ | · | ✓ | · | ✓ | · | · | · | ✓ | · | · | ✓ | · | · | ✓ | · | · |
| 67 | 6 | ✓ | · | · | ✓ | · | · | · | ✓ | · | · | · | ✓ | · | ✓ | ✓ | · | · | ✓ | · | · |
| 68 | 6 | ✓ | · | ✓ | ✓ | · | · | · | ✓ | ✓ | · | · | · | · | ✓ | · | · | · | · | · | ✓ |
| 69 | 6 | ✓ | · | · | ✓ | · | · | · | ✓ | · | · | · | ✓ | · | · | ✓ | · | · | ✓ | · | ✓ |
| 70 | 6 | ✓ | · | ✓ | ✓ | · | ✓ | · | ✓ | · | · | · | · | · | ✓ | · | ✓ | ✓ | · | · | · |
| 71 | 6 | ✓ | · | ✓ | ✓ | · | ✓ | ✓ | · | · | · | · | · | · | · | ✓ | · | ✓ | · | · | · |
| 72 | 6 | ✓ | · | ✓ | ✓ | · | ✓ | · | · | · | · | · | · | · | ✓ | ✓ | ✓ | ✓ | · | · | · |
| 73 | 6 | ✓ | · | ✓ | ✓ | · | · | · | ✓ | · | · | · | · | · | ✓ | · | ✓ | ✓ | · | · | ✓ |
| 74 | 6 | ✓ | · | ✓ | ✓ | · | · | ✓ | ✓ | · | · | · | · | · | · | · | ✓ | ✓ | · | · | ✓ |
| 75 | 6 | ✓ | · | ✓ | ✓ | · | · | · | ✓ | · | · | · | · | · | ✓ | · | ✓ | ✓ | · | · | ✓ |
| 76 | 6 | ✓ | · | ✓ | ✓ | · | ✓ | ✓ | · | · | · | · | · | · | · | ✓ | ✓ | ✓ | · | · | ✓ |
| 77 | 6 | ✓ | · | ✓ | ✓ | · | ✓ | · | · | · | · | · | · | · | ✓ | ✓ | · | ✓ | ✓ | · | · |
| 78 | 7 | ✓ | · | · | ✓ | · | · | ✓ | · | · | · | ✓ | · | · | · | · | ✓ | · | ✓ | · | · |
| 79 | 7 | ✓ | · | · | ✓ | · | · | · | ✓ | · | · | · | ✓ | · | · | ✓ | · | · | ✓ | · | · |
| 80 | 8 | ✓ | · | ✓ | ✓ | · | · | ✓ | ✓ | · | · | · | · | · | · | · | ✓ | ✓ | ✓ | · | · |
| 81 | 8 | ✓ | · | ✓ | ✓ | · | · | ✓ | ✓ | · | · | · | · | · | · | ✓ | · | ✓ | ✓ | · | · |
| 82 | 8 | ✓ | · | ✓ | ✓ | · | ✓ | ✓ | · | · | · | · | · | · | ✓ | · | ✓ | ✓ | ✓ | · | · |
| 83 | 8 | ✓ | · | ✓ | ✓ | · | ✓ | · | ✓ | · | · | · | · | · | ✓ | · | ✓ | ✓ | ✓ | · | · |
| 84 | 8 | ✓ | · | ✓ | ✓ | · | · | · | ✓ | · | · | · | · | · | ✓ | · | ✓ | · | ✓ | ✓ | · | ✓ |

Table 1: Results of our automated and exhaustive theorem search on a space of 20 axioms (including orders).

Observing these results, we first note that impossibilities can occur from all of the tested domain sizes larger than 2 onwards. This is novel in its own right since until now only impossibilities with $|X| \geq k$, $k \in \{3, 4, 6\}$ had been known.

The results themselves differ much in their level of appeal and interestingness. We can find impossibilities of at least five (potentially overlapping) categories, namely known results, variations of known results, direct consequences of other results, straightforward results, and, most importantly, new results.

Some previously known results we can easily recognise among the ones in our list: the Kannai-Peleg Theorem corresponds to our Impossibility No. 57; an impossibility theorem by Barberà and Pattanaik (1984) can be found as Impossibility No. 1. Other than that we are only aware of one more known impossibility under the interpretation of *complete uncertainty*, which we could unfortunately not encode in our framework since it uses the axiom of *neutrality* (see Section 3.1). It is a variant of the Kannai-Peleg Theorem presented by Barberà et al. (2004), in which the number of elements has been lowered to four by adding the aforementioned axiom.

Variations of known results are also easy to spot by just keeping some axioms fixed and browsing for results involving these. Impossibilities No. 80 and No. 10, for instance, are variations of the Kannai-Peleg Theorem, where in the former a weakening of the axioms makes the impossibility occur only at a larger domain size. The latter is a variation in the other direction: the additional axiom (`SUAv`) causes an impossibility at a domain size of 4 elements already. And many more such variations of known theorems can be found (e.g., No. 33, 37, 40, etc.).

As we used a set of axioms in which certain axioms imply others, we had to expect results that are just direct consequences of others. In particular, every result involving some (weak) form of independence will also occur with the standard or strict independence only, and similarly for simple dominance, which is a weaker form of the Gärdenfors principle. Examples of such results are the Impossibilities No. 3 (implied by No. 1) and Impossibility No. 9 (implied by No. 28).

Straightforward results we could only find one: Impossibility No. 2 says that a binary relation cannot fulfill both (`SUAv`) and (`SUAp`), which reflect the contradictory principles of uncertainty aversion and uncertainty appeal. This is immediate (especially when examining the exact statement of the axioms).

What we are left with are the new, i.e., previously unknown, results. There are quite a few of them, but they differ in how interesting they are. For instance, it is not very reasonable to only postulate (`GF1`) but not (`GF2`), which makes the new Impossibility No. 11 not so fascinating after all. But we can also find results like Impossibilities No. 52 or No. 56, where the combination of axioms appears to be reasonable and yet leads to an impossibility.

We will return to some of these results below. But let us now for a moment shift our perspective from problems, i.e., combinations of axioms, to the special role of individual axioms with respect to *all* results. On the one hand, the axiom (`LIN`$_\varepsilon$) of a linear order $\dot{\geq}$ on $X$ occurs in all impossibilities. This means that there is no impossibility without this axiom (on the given axiom space and up to domain size 8). This could have been anticipated: if we use the empty relation on $X$ for $\dot{\geq}$, then most axioms do not say anything about $\succeq$ anymore and hence cannot be incompatible. Also note that the only impossibility without any form of independence is the (straightforward) Impossibility No. 2. On the other hand, the axioms (evenExt) and (REFL$_\sigma$) do not occur in any impossibility. Therefore, we can conclude that these must be particularly well-compatible with the other axioms. Or put differently, that adding them to a given set of axioms does not cause an impossibility. The axiom (intIND) of intermediate independence is contained in all discovered impossibilities at domain sizes 7 and 8 (and does not cause any impossibilities at sizes 5 and below). That this axiom is involved in somewhat larger instances makes a great deal of sense intuitively: for each application of the axiom we have to add *two* elements (one above, one below the set we apply it to), and so it was to be expected that larger domain sizes are necessary for a contradiction.

In the following we discuss some of the obtained impossibilities and also provide an example of a manual proof. That we were able to quickly construct manual proofs for all theorems discussed below underpins the usefulness of our theorem search as a heuristic, even for a sceptic who may not be willing to accept the output of a SAT solver as a rigorous proof.[13] Knowing the impossible axiom sets and critical domain sizes beforehand simplified the construction of the manual proofs significantly. Additionally, one can run the search program again with slightly modified axioms, not only to get an even better understanding about where the borderline lies between the possible and the impossible, but also to have some assistance in choosing the right steps when proving the results by hand. And there is even one more application of our program when searching for a manual proof: one can run it on instances with single axioms left out and inspect the orders satisfying the remaining axioms in order to understand which structural properties these imply.

### 5.2.1 An Unintuitive Impossibility

Let us start with our most striking result. Theorem 3 of an important paper by Bossert et al. (2000) states that the axioms (SDom), (IND), (SUAv) and (STopMon) characterise the so-called *min-max ordering*, which is defined by

$$A \succeq_{mnx} B \iff \left[\min(A) \mathrel{\dot{>}} \min(B) \vee \left(\min(A) = \min(B) \wedge \max(A) \mathrel{\dot{\geq}} \max(B)\right)\right].$$

The same theorem also covers a dual result for the *max-min ordering* (characterized by the axioms (SDom), (IND), (SUAp), (SBotMon)).

The reader can now check that this contradicts the results of our theorem search since both of these axiom sets are among the impossibility theorems in Table 1 (Impossibilities No. 16 and 19). Indeed, it turns out that the proofs of Bossert et al. (2000) were flawed as Arlegi (2003) pointed out three years later. Arlegi, however, only notes that the *min-max* and *max-min orderings* do not satisfy the axiom of independence (IND), i.e., that these orders cannot be characterized by the axioms (SDom), (IND), (SUAv), (STopMon), and (SDom), (IND), (SUAp), (SBotMon), respectively. This shows the unintuitiveness of our findings (as the contrary was believed for some time), which yield more than just a counterexample to the original publication: we additionally get that the four axioms under consideration are inconsistent (in the presence of transitivity) and hence *no* transitive binary relation whatsoever can satisfy them. We now give a manual proof for this result.

---

13. We only included one example of a manual proof here. The complete set is given by Geist (2010).

**Theorem 3** (Impossibility No. 16). *Let $X$ be a linearly ordered set with $|X| \geq 4$. Then there exists no transitive binary relation $\succeq$ on $\mathcal{X}$ satisfying simple dominance (SDom), independence (IND), simple uncertainty aversion (SUAv), and simple top monotonicity (STopMon).*

*Proof.* Let $x_i$, $i \in \{1, 2, 3, 4\}$ denote four distinct elements of $X$ such that are ordered by $\dot{>}$ with respect to their index, i.e., $x_1 \dot{>} x_2 \dot{>} x_3 \dot{>} x_4$. By way of contradiction, suppose there exists a transitive binary relation $\succeq$ on $\mathcal{X}$ satisfying simple dominance (SDom), independence (IND), simple uncertainty aversion (SUAv), and simple top monotonicity (STopMon).

On the one hand, it follows from simple uncertainty aversion applied to $x_1 \dot{>} x_2 \dot{>} x_3$ that $\{x_2\} \succ \{x_1, x_3\}$, and adding $x_4$ to both sets yields (by independence):

$$\{x_2, x_4\} \succeq \{x_1, x_3, x_4\}. \tag{5}$$

On the other hand, we can use simple dominance (applied to $x_3 \dot{>} x_4$) to show $\{x_3, x_4\} \succ \{x_4\}$, from which

$$\{x_1, x_3, x_4\} \succeq \{x_1, x_4\} \tag{6}$$

follows by independence. Furthermore, simple top monotonicity applied to $x_1 \dot{>} x_2 \dot{>} x_4$ directly gives $\{x_1, x_4\} \succ \{x_2, x_4\}$, which we are able to combine with (6) by transitivity. We thus obtain

$$\{x_1, x_3, x_4\} \succ \{x_2, x_4\},$$

which directly contradicts (5). □

Note that all four axioms are not only used in the proof above, but they are *necessary* for the result and also logically independent from each other as the following automatically constructed examples of weak orders show (let $X = \{x_1, x_2, x_3, x_4\}$ and $x_1 \dot{>} x_2 \dot{>} x_3 \dot{>} x_4$):

1. The weak order $\succeq$ given by
   $\{x_1\} \succ \{x_2\} \succ \{x_3\} \succ \{x_4\} \succ \{x_1, x_2\} \succ \{x_1, x_3\} \succ \{x_2, x_3\} \succ \{x_1, x_4\} \succ \{x_2, x_4\} \succ \{x_3, x_4\} \succ \{x_1, x_2, x_3\} \succ \{x_1, x_2, x_4\} \succ \{x_1, x_3, x_4\} \succ \{x_2, x_3, x_4\} \succ \{x_1, x_2, x_3, x_4\}$
   satisfies (IND), (SUAv), (STopMon), but not (SDom).

2. The weak order $\succeq$ given by
   $\{x_1\} \succ \{x_1, x_2\} \succ \{x_2\} \succ \{x_1, x_3\} \succ \{x_2, x_3\} \succ \{x_3\} \succ \{x_1, x_2, x_3\} \succ \{x_1, x_4\} \succ \{x_2, x_4\} \succ \{x_1, x_2, x_4\} \succ \{x_3, x_4\} \succ \{x_4\} \succ \{x_1, x_3, x_4\} \succ \{x_2, x_3, x_4\} \succ \{x_1, x_2, x_3, x_4\}$
   satisfies (SDom), (SUAv), (STopMon), but not (IND).

3. The weak order $\succeq$ given by
   $\{x_1\} \succ \{x_1, x_2\} \succ \{x_1, x_3\} \sim \{x_1, x_2, x_3\} \succ \{x_2\} \succ \{x_2, x_3\} \succ \{x_3\} \succ \{x_1, x_4\} \sim \{x_1, x_2, x_4\} \sim \{x_1, x_3, x_4\} \sim \{x_1, x_2, x_3, x_4\} \succ \{x_2, x_4\} \sim \{x_2, x_3, x_4\} \succ \{x_3, x_4\} \succ \{x_4\}$
   satisfies (SDom), (IND), (STopMon), but not (SUAv).

4. The weak order $\succeq$ given by
   $\{x_1\} \succ \{x_1, x_2\} \succ \{x_2\} \succ \{x_1, x_3\} \sim \{x_1, x_2, x_3\} \succ \{x_2, x_3\} \succ \{x_3\} \succ \{x_1, x_4\} \sim \{x_1, x_3, x_4\} \sim \{x_2, x_4\} \sim \{x_1, x_2, x_4\} \sim \{x_2, x_3, x_4\} \sim \{x_1, x_2, x_3, x_4\} \succ \{x_3, x_4\} \succ \{x_4\}$
   satisfies (SDom), (IND), (SUAv), but not (STopMon).

It can now also be seen that no subset of these four axioms suffices to characterise the *min-max ordering*. Any subset containing (IND) can be rejected immediately since (IND) is violated by the *min-max ordering* (as we have noted earlier), and any subset not containing it cannot suffice for a characterisation either, since example 2 differs from the *min-max ordering* $\succeq_{mnx}$ (in which $\{x_2, x_3, x_4\} \prec_{mnx} \{x_1, x_2, x_3, x_4\}$).

Finally, we emphasise the fact that neither reflexivity nor completeness of $\succeq$ are used in the proof of Theorem 3 (as also indicated by Table 1). Thus, the impossibility already holds for arbitrary transitive binary relations instead of weak orders.

### 5.2.2 Variations of the Kannai-Peleg Theorem

Impossibility No. 9 offers an interesting variation of the Kannai-Peleg Theorem that trades an additional axiom (simple uncertainty aversion) for the impossibility occurring at a domain size of 4 rather than 6 elements.

**Theorem 4** (Impossibility No. 9). *Let $X$ be a linearly ordered set with $|X| \geq 4$. Then there exists no transitive binary relation $\succeq$ on $\mathcal{X}$ satisfying the Gärdenfors principle (GF), independence (IND) and simple uncertainty aversion (SUAv).*

The same impossibility result also holds with simple uncertainty appeal (SUAp) in place of simple uncertainty aversion (SUAv); this is Impossibility No. 12.

When we have an even closer look at Table 1, then we can see that there is an even stronger form of Theorem 4: Impossibility No. 28 corresponds to the the axioms (GF), (SUAv), (botIND), (topIND) and is impossible from domain size 4 on. In contrast to (IND), the axioms (botIND) and (topIND) allow the principle of independence in certain situations only: the element to be added has to be ranked below or above all the elements in both sets, respectively.[14] Therefore, we immediately have the following stronger version of Theorem 4.

**Theorem 5** (Impossibility No. 28). *Let $X$ be a linearly ordered set with $|X| \geq 4$. Then there exists no transitive binary relation $\succeq$ on $\mathcal{X}$ satisfying the Gärdenfors principle (GF), bottom (botIND) as well as top independence (topIND), and simple uncertainty aversion (SUAv).*

An interesting insight can be obtained from comparing Impossibility No. 48 to the previous result. It shows us that we can drop the second Gärdenfors axiom if we add just one element to the domain, i.e., we have $|X| \geq 5$. The exact result is the following:

**Theorem 6** (Impossibility No. 48). *Let $X$ be a linearly ordered set with $|X| \geq 5$. Then there exists no transitive binary relation $\succeq$ on $\mathcal{X}$ satisfying the first axiom of the Gärdenfors principle (GF1), bottom (botIND) as well as top independence (topIND), and simple uncertainty aversion (SUAv).*

Alternatively, we could have replaced (botIND) by (disIND) (No. 51), or (topIND) by (intIND), then, however, requiring at least seven elements in the domain (No. 78).

Two further variants of the Kannai-Peleg Theorem can be found in Impossibilities No. 80 and 81, which can be considered strengthenings of the original theorem as they contain weaker versions of independence only. The strengthening, however, comes at the cost of the impossibility starting from a domain size of eight elements instead of six. The form of independence that remains is a combination of intermediate, disjoint, and bottom or top independence, respectively, which (even together) are weaker than standard independence.

---

14. Actually, already in the original proof of the Kannai-Peleg Theorem (Kannai & Peleg, 1984) only these weaker forms of (IND) are used (cf. also Impossibility No. 61).

### 5.2.3 Impossibilities without Dominance

All existing impossibilities in the literature we are aware of involve the Gärdenfors principle (GF) or at least simple dominance (SDom). So let us now consider what kinds of results we can obtain without any dominance principle.

A striking impossibility without any principle of dominance —i.e., without either (GF) or (SDom)— is Impossibility No. 52: the axioms of strict independence (strictIND), simple uncertainty aversion (SUAv), and monotone consistency (MC) are incompatible in the presence of completeness and transitivity from domain size 5 on.

**Theorem 7** (Impossibility No. 52). *Let $X$ be a linearly ordered set with $|X| \geq 5$. Then there exists no weak order $\succeq$ on $\mathcal{X}$ satisfying strict independence* (strictIND), *simple uncertainty aversion* (SUAv) *and monotone consistency* (MC).

One might be tempted to think that this impossibility is mostly due to problems between (SUAv) and (MC) since they seem to express contrary ideas: whereas (SUAv) favours small sets over large ones, (MC) tells us that unions of two sets should be preferred to at least one of the sets. But actually there is even a characterisation result of the *min-max ordering* by Arlegi (2003) involving both axioms (SUAv) and (MC), demonstrating that this natural ordering fulfils these two axioms. Therefore, we can see that it should not at all be considered unreasonable to have both axioms act together.

We have found quite a few variants of this impossibility. According to our results, completeness could be replaced by simple bottom monotonicity (Impossibility No. 53) or even be dropped at the price of having one more element in the domain (Impossibility No. 56). Alternatively, one can weaken strict independence to either bottom or disjoint independence and shrink the domain by one element, at the price of adding the axiom of simple top monotonicity (Impossibilities No. 26 and 27, respectively). A seemingly further variant can be obtained from trading the axiom of extension (EXT) for a smaller domain. It is, however, a direct consequence of Impossibilities No. 26 and 27, respectively, since (EXT) and (strictIND) together imply (STopMon).

Since strict independence can be considered a relatively strong axiom, Impossibility No. 26 (and the corresponding No. 27) is worth emphasising as well, as it does only postulate a very weak form of independence.

**Theorem 8** (Impossibility No. 26). *Let $X$ be a linearly ordered set with $|X| \geq 4$. Then there exists no transitive binary relation $\succeq$ on $\mathcal{X}$ satisfying bottom independence* (botIND), *simple uncertainty aversion* (SUAv), *simple top monotonicity* (STopMon) *and monotone consistency* (MC).

This result comes as quite a surprise since Arlegi (2003) characterises the *min-max ordering* by an axiom set including (SUAv), (STopMon), and (MC) (as well as two further axioms). It follows that adding just a tiny bit of independence to these three axioms turns their possibility into a general impossibility.

## 6. Conclusion

We have presented a method for automatically verifying and discovering theorems in a subarea of economic theory concerned with the problem of formulating principles for lifting preferences over individual objects to preferences over nonempty sets of those objects. The theorems in question are impossibility theorems that establish that certain combinations of those principles (called axioms) are inconsistent. Our method has three components:

- A general result, the *universal reduction step* (a corollary of our *Preservation Theorem*), that shows that if a combination of axioms, meeting certain conditions, is inconsistent for a fixed domain size $n$, then it is also inconsistent for any domain with more than $n$ objects. The conditions on axioms for the applicability of this result are purely syntactic: any axiom that is (equivalent to) an *existentially set-guarded* (ESG) sentence in the *many-sorted logic for set preferences* (MSLSP) qualifies.

- A method for translating axioms into *propositional formulas* in CNF, and a method for *instantiating* those axioms on a computer for a fixed domain size. This allows us to *verify* small instances of an impossibility theorem using a SAT solver. Together with the universal reduction step, this then constitutes a proof of the respective impossibility theorem also for all larger domain sizes.

- A *scheduling algorithm* to search a large space of axiom combinations for different domain sizes. This, finally, allows us to systematically search for and *discover* new impossibility theorems.

We have applied our method to a set of 20 axioms that have been proposed in the literature as a means of formalising various principles for ranking sets of objects when those sets are interpreted as representing mutually exclusive alternatives from which an object will be selected in a manner that cannot be influenced by the decision maker (so-called *complete uncertainty*). This did yield a total of 84 (minimal) impossibility theorems, including both known results and new theorems. We have commented on the most interesting of these in the previous section. These results clearly demonstrate the power of our method.

This work can be extended in a number of ways. First, our method can be applied to other sets of axioms (including axioms for order types other than linear and weak orders). Implementing further axioms can be done quickly, and as long as they are covered by our universal reduction step, results can be read off after a short computation. Especially for *opportunity sets*, for which to our knowledge no impossibility results are known, the potential for success is very high.

Second, the method itself and its implementation can be refined further. It would be attractive to integrate a parser that can read our language MSLSP so that axioms no longer have to be transformed and coded by hand. A further idea is to implement dependencies between the axioms. This would make sure that only absolutely minimal results are returned, whereas now some of our results are trivial consequences of others (since some of our axioms are immediately implied by others).

Third, in case a particular combination of axioms does *not* lead to an impossibility, it may be possible to use the output of the SAT solver to infer useful information about the class of set preference orderings satisfying those axioms. Some preliminary steps in this direction have already been taken (Geist, 2010).

Finally, and this is our most tentative suggestion, it would be interesting to explore to what extent our method can be adapted to different disciplines and problem domains. A starting point might be our Preservation Theorem, which potentially can still be strengthened to a larger class of axioms. One could try to find out where the exact borderline lies between formulas that are preserved in certain substructures and those that are not. For arbitrary first-order models this has been done in the famous Łoś-Tarski Theorem, but for our class of structures for set preferences it is still an open question.

## Acknowledgments

We would like to thank Umberto Grandi and three anonymous JAIR reviewers for a host of helpful comments and suggestions on earlier versions of this paper.

## Appendix A. List of Axioms

In this appendix we provide the complete list of the axioms used in the theorem search presented in Section 5. These axioms (or variations thereof) and further references can all be found in the survey by Barberà et al. (2004).

The first axioms given here are the *order axioms*. For one, there are the axioms describing the linear order $\dot{\geq}$ on $X$, for another, the ones describing a weak order $\succeq$ on $\mathcal{X} = 2^X \setminus \{\emptyset\}$. The former will just be denoted by ($\texttt{LIN}_\varepsilon$), whereas the latter are split up into their three components reflexivity ($\texttt{REFL}_\sigma$), completeness ($\texttt{COMPL}_\sigma$) and transitivity ($\texttt{TRANS}_\sigma$), which are then treated as separate axioms in order to investigate which parts are actually necessary for impossibilities. The axioms in their intuitive form are:

$$
\begin{array}{lll}
(\texttt{LIN}_\varepsilon) & x \dot{\geq} x \text{ for all } x \in X & \text{(reflexivity)} \\
& x \dot{\geq} y \vee x \dot{\leq} y \text{ for all } x \neq y \in X & \text{(completeness)} \\
& x \dot{\geq} y \wedge y \dot{\geq} z \Rightarrow x \dot{\geq} z \text{ for all } x, y, z \in X & \text{(transitivity)} \\
& x \dot{\geq} y \wedge y \dot{\geq} x \Rightarrow x = y \text{ for all } x, y \in X & \text{(antisymmetry)}
\end{array}
$$

$$
\begin{array}{lll}
(\texttt{REFL}_\sigma) & A \succeq A \text{ for all } A \in \mathcal{X} & \text{(reflexivity)} \\
(\texttt{COMPL}_\sigma) & A \succeq B \vee A \preceq B \text{ for all } A \neq B \in \mathcal{X} & \text{(completeness)} \\
(\texttt{TRANS}_\sigma) & A \succeq B \wedge B \succeq C \Rightarrow A \succeq C \text{ for all } A, B, C \in \mathcal{X} & \text{(transitivity)}
\end{array}
$$

Next we have the axiom of *extension*, which is a very natural requirement and thus also implied by some other axioms (e.g., the Gärdenfors principle):

$$(\texttt{EXT}) \quad x \dot{\geq} y \iff \{x\} \succeq \{y\} \text{ for all } x, y \in X$$

A further set of axioms we included in our search is the one dealing with the concept of *dominance*, i.e., the idea that adding an object $x$ to a set of objects $A$ that are all dominated by (or dominating) the object $x$ produces a better (or worse) set, respectively. We chose for the well-known *Gärdenfors principle* (`GF`), which was introduced in Section 2.2 already, as well as a weaker version by Barberà (1977) called *simple dominance* (`SDom`), which restricts (`GF`) to small sets:

$$\begin{aligned}
&\text{(GF1)} && ((\forall a \in A) x \mathrel{\dot{>}} a) \Rightarrow A \cup \{x\} \succ A \text{ for all } x \in X \text{ and } A \in \mathcal{X} \\
&\text{(GF2)} && ((\forall a \in A) x \mathrel{\dot{<}} a) \Rightarrow A \cup \{x\} \prec A \text{ for all } x \in X \text{ and } A \in \mathcal{X} \\
&\text{(SDom)} && x \mathrel{\dot{>}} y \Rightarrow (\{x\} \succ \{x,y\} \land \{x,y\} \succ \{y\}) \text{ for all } x,y \in X
\end{aligned}$$

Independence axioms are also commonly postulated and especially their weaker variants or versions thereof, like bottom, top, disjoint and intermediate independence, frequently play a role in characterisation results (e.g., see Pattanaik & Peleg, 1984; Nitzan & Pattanaik, 1984). We decided to include standard independence (as already introduced in Section 2.2), a stronger version (`strictIND`), which implies *strict* preferences, and a few weaker versions, viz. bottom (`botIND`), top (`topIND`), disjoint (`disIND`) and intermediate independence (`intIND`), which only apply to certain combinations of sets and elements.

$$\begin{aligned}
&\text{(IND)} && A \succ B \Rightarrow A \cup \{x\} \succeq B \cup \{x\} \text{ for all } A, B \in \mathcal{X} \text{ and } x \in X \setminus (A \cup B) \\
&\text{(strictIND)} && A \succ B \Rightarrow A \cup \{x\} \succ B \cup \{x\} \text{ for all } A, B \in \mathcal{X} \text{ and } x \in X \setminus (A \cup B) \\
&\text{(botIND)} && A \succ B \Rightarrow A \cup \{x\} \succeq B \cup \{x\} \text{ for all } A, B \in \mathcal{X} \\
& && \text{and } x \in X \setminus (A \cup B) \text{ such that } y \mathrel{\dot{>}} x \text{ for all } y \in A \cup B \\
&\text{(topIND)} && A \succ B \Rightarrow A \cup \{x\} \succeq B \cup \{x\} \text{ for all } A, B \in \mathcal{X} \\
& && \text{and } x \in X \setminus (A \cup B) \text{ such that } x \mathrel{\dot{>}} y \text{ for all } y \in A \cup B \\
&\text{(disIND)} && A \succ B \Rightarrow A \cup \{x\} \succeq B \cup \{x\} \text{ for all } A, B \in \mathcal{X}, \\
& && \text{such that } A \cap B = \emptyset, \text{ and for all } x \in X \setminus (A \cup B) \\
&\text{(intIND)} && A \succ B \Rightarrow A \cup \{x,y\} \succeq B \cup \{x,y\} \text{ for all } A, B \in \mathcal{X} \text{ and } x, y \in X \setminus (A \cup B) \\
& && \text{such that } x \mathrel{\dot{>}} z \text{ and } z \mathrel{\dot{>}} y \text{ for all } z \in A \cup B
\end{aligned}$$

Bossert (1997) introduced axioms describing the attitude of the decision maker towards uncertainty. We formalise weakenings of these axioms that apply to small sets only, since these are sufficient for characterisation results like those of Arlegi (2003). *Uncertainty aversion* postulates that the decision maker will, for any alternative $x$, (strictly) prefer this alternative to a set containing both a better and a worse alternative. *Uncertainty appeal*, on the other hand, says that the ranking has to be just the other way around: the set with a better and a worse element is (strictly) preferred to the single element $x$.

$$\begin{aligned}
&\text{(SUAv)} && (x \mathrel{\dot{>}} y \land y \mathrel{\dot{>}} z) \Rightarrow \{y\} \succ \{x,z\} \text{ for all } x,y,z \in X \\
&\text{(SUAp)} && (x \mathrel{\dot{>}} y \land y \mathrel{\dot{>}} z) \Rightarrow \{x,z\} \succ \{y\} \text{ for all } x,y,z \in X
\end{aligned}$$

Arlegi (2003) also uses two monotonicity axioms, called *simple top* and *bottom monotonicity*. The underlying idea is simple: given two alternatives, it is better to get the better one of the two together with some third element (instead of the worse one with the same third element). The two variants of the axiom then only apply to alternatives that are ranked higher (top) than the third alternative, or ranked lower (bottom), respectively.

$$\begin{aligned}
&\text{(STopMon)} && x \mathrel{\dot{>}} y \Rightarrow \{x,z\} \succ \{y,z\} \text{ for all } x,y,z \in X \text{ such that } x \mathrel{\dot{>}} z \text{ and } y \mathrel{\dot{>}} z \\
&\text{(SBotMon)} && y \mathrel{\dot{>}} z \Rightarrow \{x,y\} \succ \{x,z\} \text{ for all } x,y,z \in X \text{ such that } x \mathrel{\dot{>}} y \text{ and } x \mathrel{\dot{>}} z
\end{aligned}$$

A rather odd axiom is the principle of *even-numbered extension of equivalence*. It says that, for all sets with an even number of elements, if the decision maker is indifferent about whether this set is added to each of two distinct singleton sets, then she should also be indifferent about whether it is added to the union of the two singleton sets. Even though it lacks intuitive support, this axiom is useful because (together with a few other principles) it characterises a median-based ordering proposed by Nitzan and Pattanaik (1984).

$$(\texttt{evenExt}) \quad (A \cup \{x\} \sim \{x\} \wedge A \cup \{y\} \sim \{y\}) \Rightarrow A \cup \{x,y\} \sim \{x,y\}$$
$$\text{for all } A \in \mathcal{X}, \text{ such that } |A| \text{ is even, and for all } x,y \in X \setminus A$$

The final axiom in our list is *monotone consistency* (MC), which was put forward by Arlegi (2003) to characterise (in connection with other axioms) the *min-max ordering* (see also Section 5.2). (MC) expresses that if a set of objects $A$ is at least as good as another set $B$, then the union of the two is at least as good as the latter. This implies —and for complete binary relations is equivalent to— the potentially worse set $B$ not being strictly better than the union of the two. Intuitively, it means that after adding the alternatives of the (weakly preferred) set $A$ to the set $B$, the decision maker maintains the alternatives she had in $B$ plus the ones that were contained in $A$, which was weakly preferred to $B$. Thus, this process should not produce a set that is strictly worse than $B$.

$$(\texttt{MC}) \quad A \succeq B \Rightarrow A \cup B \succeq B \text{ for all } A, B \in \mathcal{X}$$

Although (MC) appears to be similar to the first axiom of the Gärdenfors principle, it is in fact quite different since it does not dictate the existence of any *strict* preferences.

Automated Search for Impossibility Theorems in Social Choice Theory

173